\documentclass[letterpaper]{article} 
\usepackage{aaai2026}  
\usepackage{times}  
\usepackage{helvet}  
\usepackage{courier}  
\usepackage[hyphens]{url}  
\usepackage{graphicx} 

\usepackage{natbib}  
\usepackage{caption} 

\usepackage{url}            
\usepackage{amsfonts}       
\usepackage{nicefrac}       
\usepackage{microtype}      
\usepackage{xcolor}         
\usepackage{graphicx}
\usepackage{todonotes}
\usepackage{caption}
\usepackage{amsmath}
\usepackage{booktabs}
\usepackage{placeins} 

\usepackage{xcolor}
\usepackage{colortbl}         
\usepackage{listings}

\lstdefinestyle{pythonstyle}{
    language=Python,
    basicstyle=\ttfamily\small,
    keywordstyle=\color{blue},
    commentstyle=\color{gray},
    stringstyle=\color{orange},
    showstringspaces=false,
    numbers=left,
    numberstyle=\tiny\color{gray},
    stepnumber=1,
    numbersep=5pt,
    backgroundcolor=\color{white},
    frame=single,
    tabsize=4,
    captionpos=b
}

\definecolor{darkred}{RGB}{139, 0, 0} 
\definecolor{darkblue}{RGB}{0, 0, 139} 

\definecolor{yellow}{rgb}{1, 1, 0.7}
\definecolor{orange}{rgb}{1, 0.85, 0.7}
\definecolor{tablered}{rgb}{1, 0.7, 0.7}
\definecolor{red}{rgb}{1, 0, 0}

\usepackage{algorithm}
\usepackage{algorithmic}

\usepackage{newfloat}
\usepackage{listings}
\DeclareCaptionStyle{ruled}{labelfont=normalfont,labelsep=colon,strut=off} 
\floatstyle{ruled}
\newfloat{listing}{tb}{lst}{}
\floatname{listing}{Listing}
\definecolor{navyblue}{rgb}{0.0, 0.0, 0.5}
\usepackage[
    colorlinks=true, 
    urlcolor=cyan, 
    linkcolor=blue, 
    citecolor=navyblue
]{hyperref}

\title{EarthCrafter: Scalable 3D Earth Generation via Dual-Sparse Latent Diffusion}

\author{%
  Shang Liu$^{1,2}$*, Chenjie Cao$^{1,2,3}$*, Chaohui Yu$^{1,2}$\textsuperscript{\dag}, Wen Qian$^{1,2}$, Jing Wang$^{1,2}$, Fan Wang$^{1}$\\
  $^1$DAMO Academy, Alibaba Group, 
  $^2$Hupan Lab,
  $^3$Fudan University \\
  {\tt\small \{liushang.ls,caochenjie.ccj,huakun.ych, qianwen.qian, yunfei.wj, fan.w\}@alibaba-inc.com\\}
  \small{Project page (code, model, and data): \url{https://whiteinblue.github.io/earthcrafter/}}
  \vspace{-0.2in}
}

\usepackage{bibentry}

\begin{document}

\makeatletter
\let\@oldmaketitle\@maketitle
\renewcommand{\@maketitle}{\@oldmaketitle
 \centering
    \vspace{-.4in}
    \includegraphics[width=0.95\linewidth]{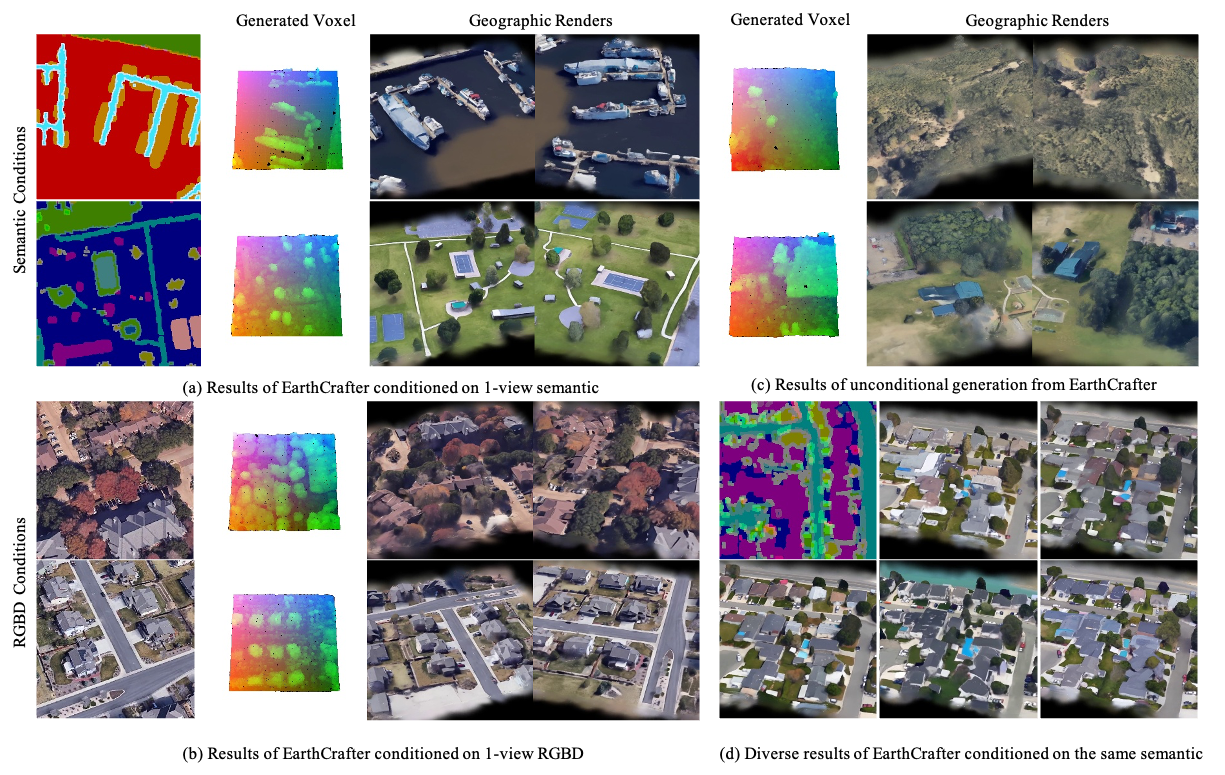}
    \vspace{-.1in}
    \captionof{figure}{EarthCrafter enjoys impressive generations conditioned on various guidance, including (a) 1-view aerial semantic and (b) 1-view RGBD. (c) EarthCrafter is powerful enough to handle unconditional generation, sampling reasonable geographic-scale 3D assets from the prior distribution. (d) EarthCrafter is enabled to produce diverse outcomes.}
    \label{fig:teaser}
  \bigskip}
\makeatother

\maketitle

\begin{abstract}
Despite the remarkable developments achieved by recent 3D generation works, scaling these methods to geographic extents, such as modeling thousands of square kilometers of Earth’s surface, remains an open challenge.
We address this through a dual innovation in data infrastructure and model architecture.
First, we introduce Aerial-Earth3D, the largest 3D aerial dataset to date, consisting of 50k curated scenes (each measuring 600m$\times$600m) captured across the U.S. mainland, comprising 45M multi-view Google Earth frames.
Each scene provides pose-annotated multi-view images, depth maps, normals, semantic segmentation, and camera poses, with explicit quality control to ensure terrain diversity.
Building on this foundation, we propose EarthCrafter, a tailored framework for large-scale 3D Earth generation via sparse-decoupled latent diffusion. Our architecture separates structural and textural generation:
1) Dual sparse 3D-VAEs compress high-resolution geometric voxels and textural 2D Gaussian Splats (2DGS) into compact latent spaces, largely alleviating the costly computation suffering from vast geographic scales while preserving critical information. 
2) We propose condition-aware flow matching models trained on mixed inputs (semantics, images, or neither) to flexibly model latent geometry and texture features independently.
Extensive experiments demonstrate that EarthCrafter performs substantially better in extremely large-scale generation.
The framework further supports versatile applications, from semantic-guided urban layout generation to unconditional terrain synthesis, while maintaining geographic plausibility through our rich data priors from Aerial-Earth3D.
\end{abstract}


\section{Introduction}

The field of 3D generation has witnessed remarkable progress in recent years, evolving from object-level~\cite{liu2023zero,shi2023zero123plus,liu2024one,xiang2024structured,ren2024xcube,zhao2025hunyuan3d} to scene-level~\cite{ren2024scube,zhou2024dreamscene360,gao2024cat3d,li2024director3d,yang2024prometheus,zhang2025flare} synthesis, yielding impressive photorealistic and structurally coherent outcomes.
Moreover, recent works have pushed these capabilities toward urban-scale generation under diverse conditions~\cite{xie2024citydreamer,xie2024gaussiancity,deng2024citycraft,engstler2025syncity}.
These achievements lead to new applications in computer graphics, virtual reality, and high-fidelity geospatial modeling.

Despite these achievements, a critical gap remains in scaling 3D generation to extensive geographic-scale—\emph{a domain requiring holistic modeling of both anthropogenic structures and natural terrains.} We identify two fundamental limitations in existing approaches:
1) Most urban generation frameworks solely focus on the city generation within constrained semantic scopes~\cite{xie2024gaussiancity,deng2024citycraft,engstler2025syncity}, neglecting other diverse natural formations (\textit{e.g.}, mountains, lakes, and deserts). 
This requires comprehensive aerial datasets encompassing multi-terrain formations and well-designed models containing scalable capacity to handle the general Earth generation.
2) Since the large-scale 3D generation is inherently intractable, existing generative methods heavily depend on various conditions, including images, semantics, height fields, captions, or combinations of them~\cite{xie2024citydreamer, xie2025citydreamer4d, shang2024urbanworld, yang2024prometheus, ren2024scube, xiang2024structured}. 
While these conditions improve the results, they constrain generative flexibility. 
Conversely, unconditional generation at geographic scales often collapses into geometric incoherence or textural ambiguity, failing to produce satisfactory outcomes.

To address these challenges, we improve both data curation and model architecture to  enhance geographic-scale generation.
Formally, we present \textbf{Aerial-Earth3D}, the largest 3D aerial dataset created to date. This dataset comprises 50,028 meticulously curated scenes, each spanning 600m$\times$600m, sourced across the mainland U.S. with 45 million multi-view frames captured from Google Earth. 
To effectively cover valid and diverse regions with limited viewpoints, we carefully design heuristic camera poses based on simulated 3D scenes built upon DEM~\cite{dem}, OSM~\cite{openstreetmap}, and MS-Building~\cite{GlobalMLBuildingFootprints} datasets.
Since Google Earth does not provide source meshes, we reconstruct 3D meshes via InstantNGP~\cite{mueller2022instant}, applying several post-processing techniques to extract surface planes, fix normals, and refine mesh connectivity. 
Then these meshes are voxelized as the ground truth for structural generation.
Additionally, we employ AIE-SEG~\cite{xu2023analytical} to create semantic maps as mesh attributes, comprising 25 distinct classes.
As summarized in Table~\ref{tab:dataset_comparison}, Aerial-Earth3D stands out as a large-scale 3D aerial dataset characterized by its diverse terrains and 3D annotations, significantly advancing both 3D generation and reconstruction efforts.

Building upon this robust dataset, we present \textbf{EarthCrafter}, a novel framework designed for geographic-scale 3D generation through dual-sparse latent diffusion. 
Following Trellis~\cite{xiang2024structured}, 
EarthCrafter inherits the advantages of disentangled structure and texture generations with flexible conditioning and editing capabilities.
However, Trellis focuses on object-level generation rather than the geographic scene, while the latter instance contains 10 times more voxels for the geometric modeling, presenting significant challenges in feature storage efficiency, geometric compression, network design, and input condition alignment.
Thus, we propose several key innovations to extend this method to a geographic scale.
Specifically, EarthCrafter integrates dual-sparse VAEs~\cite{xiang2024structured} and Flow Matching (FM) diffusion models~\cite{esser2024scaling} for structure and texture generations, respectively. 
During the training of the texture VAE, which directly decodes 2D Gaussian Splatting (2DGS)~\cite{huang20242d} as textural representation, we find that high-resolution voxel features within low channels~\cite{flux2024} substantially outperform spatially compressed voxel features with large channels~\cite{oquab2023dinov2} in large-scale 3D generation, while the former enjoys a lighter I/O overhead.
In contrast to~\cite{xiang2024structured}, we further spatially compress voxel representations of structured VAE via elaborate sparse network design, which allows us to efficiently represent detailed geographic shapes with 97.1\% structural accuracy.
Additionally, we improve the model designs for both textual and structural FM models to tame the extremely large-scale generation.
These models can be flexibly conditioned on images, semantics, or operate without  conditions.
Especially, we employ a novel coarse-to-fine framework for structural FM, which begins by classifying the full voxel initialization into a coarse voxel space, followed by a refinement phase that converts to a fine voxel space while predicting the related latents.
This coarse-to-fine modeling enables more precise structures compared to the one-stage dense modeling.

We conduct extensive experiments to verify the effectiveness of the proposed method. The key contributions of this paper can be summarized as follows:
\begin{itemize}
    \item \textbf{Aerial-Earth3D} is presented as the largest 3D aerial dataset, comprising images captured from diverse structures and natural terrains with annotated 3D presentations.
    \item \textbf{Dual-sparse VAEs} are designed for structural and textural  encoding, facilitating efficient I/O, superior appearance, and detailed structures for large-scale generation.
    \item \textbf{Tailored flow matching models} are proposed to enhance the modeling of latent spaces, while the coarse-to-fine strategy is incorporated for precise structural generation.
\end{itemize}

\begin{table}[!t]
\centering
\vspace{-0.1in}
\vspace{-0.1in}
\scriptsize
\setlength{\tabcolsep}{3pt}
\begin{tabular}{l|ccccl}
\toprule
Dataset & Area & Images & Sites & Class & Source 
\\ \midrule
UrbanScene3D~\cite{UrbanScene3D} & 136 & 128K & 16 & 1 &  Synth/Real  \\
CityTopia~\cite{xie2025citydreamer4d} & 36 & 3.75K  & 11 & 7 & Synth \\
CityDreamer~\cite{xie2024citydreamer} & 25  & 24K  & 400  & 6  & Real \\
Building3D~\cite{wang2023building3d} & 998 & -  &  16 &  1 & Synth \\
MatrixCity~\cite{li2023matrixcity} & 28 & 519K  &  2 &  - & Synth \\
STPLS3D~\cite{chen2022stpls3d} & 17 & 62.6K  &  67 &  \textbf{32} & Synth/Real \\
SensatUrban~\cite{hu2021towards}  & 7.6 & -  &  3 &  13 & Synth/Real \\
\midrule
\textbf{Ours} & \textbf{18010} & \textbf{45M}  &  \textbf{50K} &  25 & Real  \\
\bottomrule
\end{tabular}
\captionof{table}{\textbf{Comparison of aerial-view 3D scene datasets.}}
\label{tab:dataset_comparison}
\vspace{-0.2in}
\end{table}

\section{Related Work}
\label{sec:related_work}

\paragraph{3D Generative Models.}
Recent advances in 3D generative models have garnered significant attention.
Particularly, the rise of 2D generation models~\cite{rombach2022high,esser2024scaling,flux2024} has led to an exploration of their potential for 3D-aware generation.
One line of research involves fine-tuning 2D diffusion models to allow for pose-conditioned Novel View Synthesis (NVS) in objects~\cite{liu2023zero,liu2023one,shi2023zero123plus,shi2023mvdream,wang2023imagedream} or scenes~\cite{sargent2024zeronvs,hollein2024viewdiff,wu2024reconfusion,gao2024cat3d,cao2024mvgenmaster}. Since their outcomes are primarily multi-view 2D images, converting them into high-quality 3D representations remains a challenge due to inherent view inconsistencies.
Pioneering work has also been done in distilling priors from 2D diffusion models while optimizing 3D representations~\cite{mildenhall2021nerf,kerbl20233d} via Score Distillation Sampling (SDS)~\cite{poole2023dreamfusion,kim2023collaborative,zhu2023hifa,wang2023prolificdreamer}.
However, the test-time SDS is not efficient enough and often produces inferior 3D assets due to over-saturation and multi-face Janus artifacts.
Additionally, some research has achieved the creation of large 3D scenes by iteratively stitching 3D representations within depth warping and inpainting novel views~\cite{fridman2023scenescape,hollein2023text2room,yu2024wonderjourney,liang2024luciddreamer,shriram2024realmdreamer,yu2024wonderworld}, suffering from the prohibitive inference cost of test-time dataset updates.
Despite the substantial progress made by these approaches, challenges related to 3D consistency, quality, and efficiency continue to impede their extension to complex geographic-scale 3D generation.

\paragraph{Feed-Forward 3D Generation.}
To overcome undesired error accumulation caused by NVS and costly test-time optimization of SDS and warping-inpainting pipelines, several approaches have emerged that directly predict 3D representations, including NeRF~\cite{chan2023generative,xu2023dmv3d,hong2024lrm}, 3D Gaussian Splatting (3DGS)~\cite{tang2024lgm,zou2024triplane,ren2024scube,chen2024mvsplat360,li2024director3d,yang2024prometheus}.
Following them, Trellis~\cite{xiang2024structured} enhances the capability by employing a decoupled generative pipeline that separately models structured intermediates and various types of final 3D outcomes. But Trellis struggles to generate object-level 3D assets with constrained geometric details for large-scale scenes.
Moreover, SCube~\cite{ren2024scube} advances this two-stage approach to encompass scene-level generation.
However, feed-forward 3D generation is often data-hungry. SCube only considered open-released autonomous driving datasets, lacking exploration into more challenging geographic scenarios.
Additionally, many feed-forward methods~\cite{ren2024scube,chen2024mvsplat360} stand between the 3D reconstruction and the 3D generation, heavily depending on input conditions such as sparse-view or single-view images.
This dependency significantly limits their flexibility for broader applications, including texture editing.

\section{Aerial-Earth3D Dataset}
\label{sec:dataset}



\begin{figure}
\vspace{-0.1in}
\centering
\includegraphics[width=1.0\linewidth]
{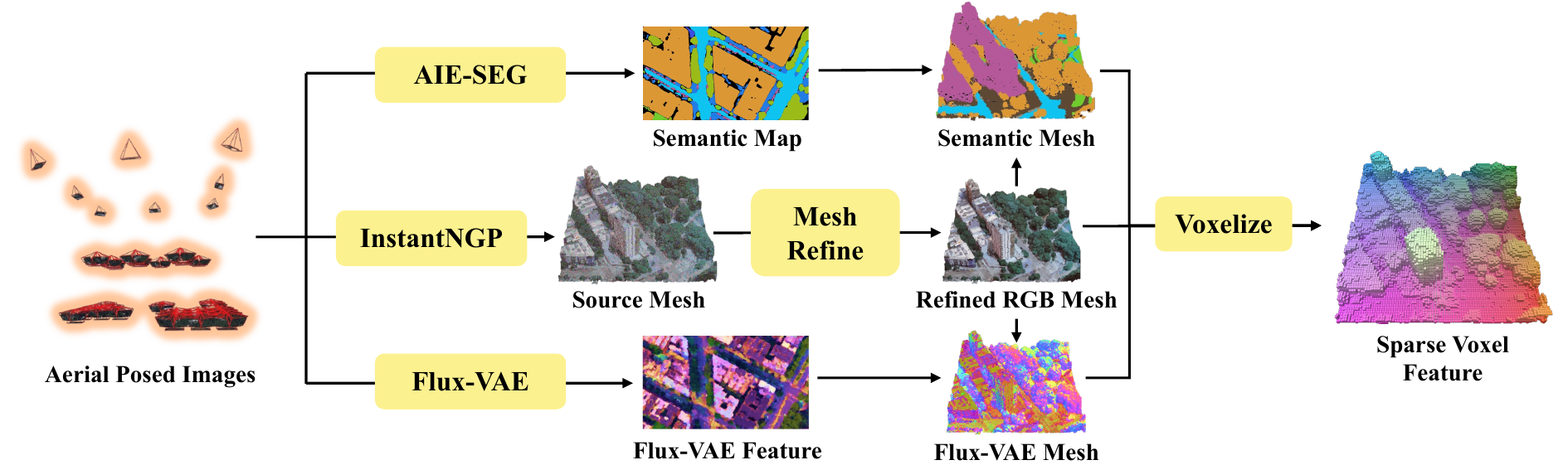}
\vspace{-0.2in}
\caption{\textbf{The overall data pipeline of Aerial-Earth3D.} InstantNGP is utilized to achieve source meshes, which are refined with heuristic strategies. Multi-view Flux-VAE features and semantic maps are aggregated on meshes. Then, these featured meshes are voxelized as inputs to TexVAE.
}
\label{fig:data_pipeline}
\vspace{-0.2in}
\end{figure}

The Aerial-Earth3D dataset was developed through a rigorous process of data curation. Initially, high-quality scenes were sampled from the ``Things to do'' recommendations in Google Earth ~\cite{EarthMap}, which yielded approximately 150,745 sites of interest across the continental United States. 
Then, we integrated the OSM driving roads~\cite{openstreetmap}, DEM terrain~\cite{dem}, and MS-Building~\cite{GlobalMLBuildingFootprints} height data to construct a high-precision simulated 3D scene, and utilized the simulated scene to design a comprehensive viewpoint planning scheme. 
Next, we feed those viewpoints to Google Earth Studio~\cite{EarthStudio} to render corresponding scene images. 
Subsequently, we utilize the InstantNGP~\cite{mueller2022instant} to reconstruct each scene and use marching cube to export 3D scene meshes, and then those meshes are refined through topological repairs.
At last, we draw attributes of color, normal, feature, and semantic to those meshes by a score aggregation algorithm. In detail, we use the AIE-SEG~\cite{xu2023analytical} model for semantic predictions and utilize the Flux-VAE~\cite{flux2024} encoder to export features; this process is shown in Figure~\ref{fig:data_pipeline}.
Ultimately, we successfully build 50,028 high-quality 3D scenes, and details are presented in the supplementary.

\section{Method}

\begin{figure*}
\centering
\includegraphics[width=1\linewidth]
{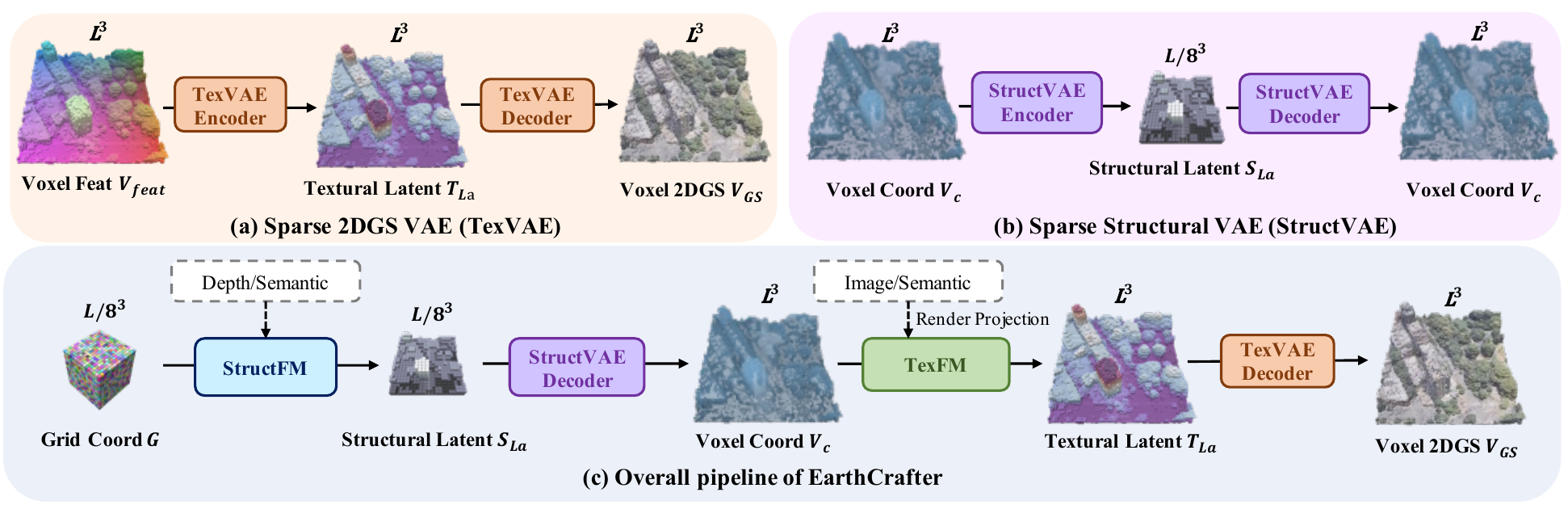}
\vspace{-0.2in}
   \caption{\textbf{Overview of EarthCrafter.}
   EarthCrafter separately models texture and structure in the latent space compressed by TexVAE and StructVAE as illustrated in (a) and (b), respectively.
   EarthCrafter also contains textural and structural flow-matching models, \textit{i.e.}, TexFM and StructFM, to model related latent presentations.
   We show the overall pipeline of EarthCrafter in (c), while dashed boxes denote optional conditions.}
\label{fig:overview}
\vspace{-0.2in}
\end{figure*}

\paragraph{Overview.}
We show the overall pipeline of EarthCrafter in Figure~\ref{fig:overview}(c), comprising separate structure and texture generations.
Given randomly initialized 3D noise grid coordinates $G\in\mathbb{R}^{(\frac{L}{8})^{3}\times{3}}$, the structural flow matching model (StructFM) generates structural latents $S_{La}\in\mathbb{R}^{(\frac{L}{8})^{3}\times c_s}$ with optional depth or semantic conditions, where $c_s$ indicates the channel of $S_{La}$. Then, the structural VAE (StructVAE) decoder is utilized to decode $S_{La}$ to high-resolution voxel coordinates $V_c\in\mathbb{R}^{L^3\times 3}$.
Subsequently, the textural flow matching model (TexFM) generates textural latent $T_{La}\in\mathbb{R}^{L^3\times c_t}$ based on $V_c$ within optional image and semantic conditions, where $c_t$ indicates the channel of $T_{La}$.
We employ the textural VAE (TexVAE) decoder to recover voxel 2DGS $V_{GS}\in\mathbb{R}^{L^3\times 16}$ as the final 3D presentation.

\subsection{Dual-Sparse VAEs}

\begin{figure}
\centering
\includegraphics[width=1.0\linewidth]
{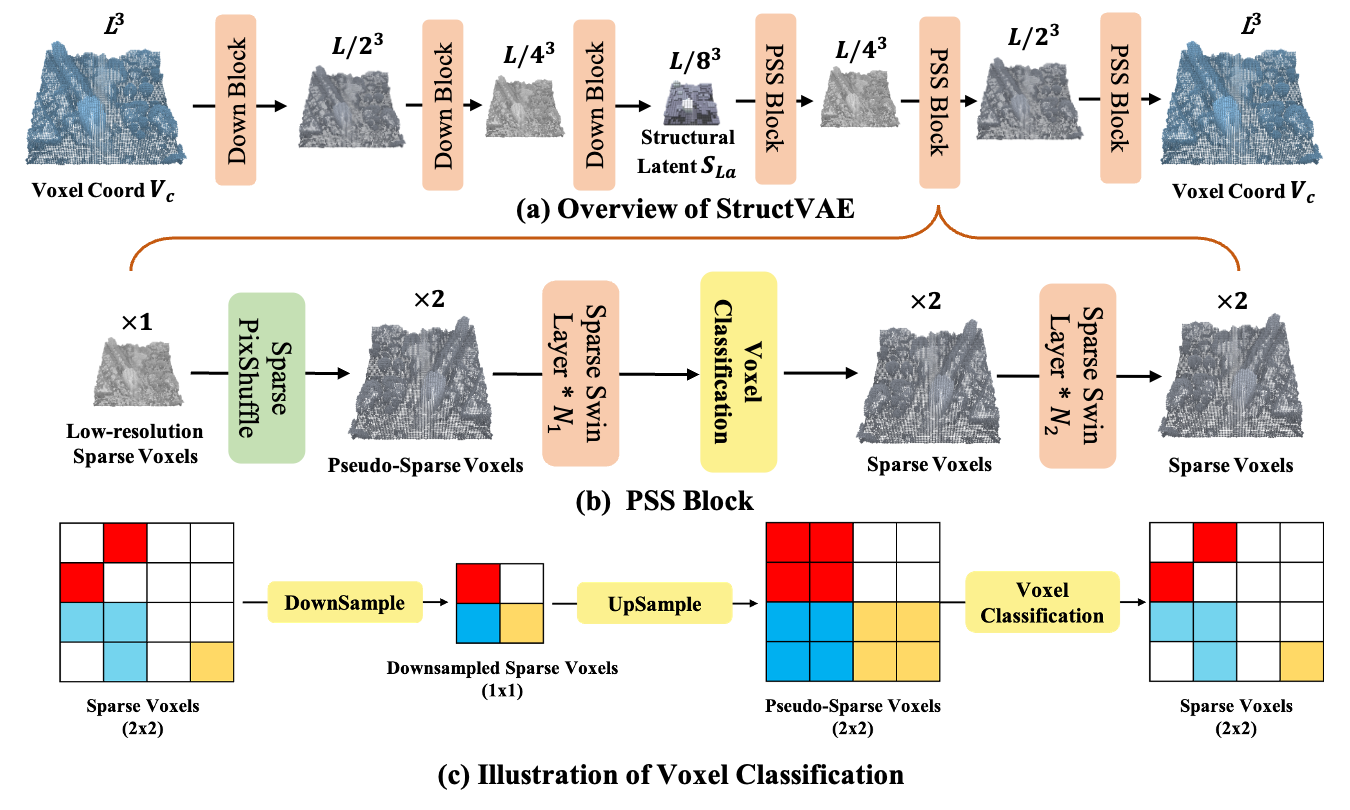}
\vspace{-0.2in}
   \caption{\textbf{StructVAE.} 
   (a) Overview of encoder-decoder based StructVAE. (b) Pseudo-Sparse to Sparse (PSS) block is used to upsample voxels and then classify them from pseudo-sparse voxels into sparse outcomes as in (c).
    \label{fig:structvae}}
\vspace{-0.2in}
\end{figure}

\subsubsection{StructVAE.}
\label{sec:structvae}
Differing from the dense architecture of Trellis~\cite{xiang2024structured}, which suffers from a fixed low-resolution voxel space, we propose StructVAE, leveraging a spatially compressed structural latent space within sparse voxel modeling to enhance efficiency.
As shown in Figure~\ref{fig:structvae}(a), StructVAE utilizes an encoder-decoder framework to compress the full voxel coordinates $V_c\in\mathbb{R}^{L^3\times 3}$ into $S_{La}\in\mathbb{R}^{(\frac{L}{8})^{3}\times c_s}$, achieving a reduction to 1/256 of the original size, where $c_s=32$.
We first incorporate positional encoding into $V_c$. 
Subsequently, 4 transformer layers and one sparse 3D convolution layer~\cite{williams2024fvdb} with a stride of 2 are applied to conduct the voxel downsampling.
We should claim that upsampling sparse voxels presents greater challenges than downsampling. Because it is non-trivial to restore accurate sparse geometry after the naive upsample strategies as shown in Figure~\ref{fig:structvae}(c).
Thus, we present the novel Pseudo-Sparse to Sparse (PSS) block to enable precise upsampling of sparse geometry via \emph{sparse pixel shuffle} and \emph{voxel classification}, as detailed in Figure~\ref{fig:structvae}(b). 
Specifically, the tailored pixel shuffle layer is designed to upsample sparse representations, with upsampled voxels termed as pseudo-sparse voxels. This definition indicates that some upsampled voxels are invalid and should be discarded to maintain accurate geometry with reasonable sparsity.
Consequently, we propose to employ a classification module to recover valid sparse voxels for each PSS block during the upsampling as shown in Figure~\ref{fig:structvae}(c).
Moreover, sparse Swin transformer~\cite{xiang2024structured} is leveraged to improve the upsample learning of PSS blocks, while full attention layers show superior capacity to learn low-resolution features with $(\frac{L}{8})^3$.
StructVAE adheres to the VAE learning objective from XCube~\cite{ren2024xcube}.
The innovative StructVAE's architecture achieves both spatially compressed geometry and 97.1\% accuracy of structural reconstruction to save the computation of the following generation.

\subsubsection{TexVAE.}
\label{sec:texvae}

\emph{1) Voxelized Features.} Following Trellis~\cite{xiang2024structured}, we aggregate features from images to inject appearance information into geometric voxels.
However, challenges persist in the large-scale learning of TexVAE.
While we have implemented spatial compression for StructVAE, applying similar feature compression in TexVAE significantly degrades texture recovery performance, as confirmed by our pilot studies. 
Additionally, learning TexVAE with high-dimensional voxelized features is I/O intolerable; for instance, utilizing 1024-d features from DINOv2~\cite{oquab2023dinov2} in Trellis~\cite{xiang2024structured} would require approximately 471M storage for each scene.
Therefore, we propose to use \textit{fine-grained, low-channel features} instead of \textit{coarse, large-channel features} for large-scale texture VAE learning.
Formally, we select the VAE features trained for FLUX~\cite{flux2024} (16-channel) as our feature extractor, significantly reducing the feature dimensionality compared to DINOv2. Although FLUX-VAE is tailored for image generation, it demonstrates impressive reconstruction capabilities for texture recovery in our study.
To further improve the presentation, we employ the hierarchical FLUX-VAE features through nearest resizing, denoted as $[f_{0};f_{1};f_{2}]\in\mathbb{R}^{16*3=48}$, where the features are concatenated at scales of 1/1, 1/2, and 1/4, respectively.
Moreover, we incorporate the cross-shaped RGB pixels $f_{rgb}\in\mathbb{R}^{5*3=15}$ and normal features $f_{n}\in\mathbb{R}^{3}$ as additional features.
The final voxelized feature can be expressed as the concatenation:
\begin{equation}
\label{eq:texvae_feat}
f_{feat} = [f_{0};f_{1};f_{2};f_{rgb};f_{n}]\in\mathbb{R}^{66},
\end{equation}
which only occupies 31M storage for each scene (6.4\% compared to DINOv2).
Following VCD-Texture~\cite{liu2024vcd}, we utilize the score-aggregation to aggregate voxelized features into $V_{feat}$ according to distances between projected pixels and voxels, as well as view scores. 

\emph{2) Model Designs and Learning Objectives.} 
The network architecture of TexVAE follows Trellis~\cite{xiang2024structured}, utilizing an encoder-decoder model enhanced with 12 sparse Swin transformer layers for each component.
The encoder of TexVAE converts the voxelized features $V_{feat}\in\mathbb{R}^{L^3\times 66}$ to textural latents $T_{La}\in\mathbb{R}^{L^3\times c_t}, c_t=8$, while the decoder is utilized to translate latent features to 2DGS presentations. 
These representations consist of offset $o_i$, scaling $s_i$, opacity $\alpha_i$, rotation matrix $R_i$, and spherical harmonics $c_i$. The loss function of TexVAE comprises L1, LPIPS~\cite{zhang2018unreasonable}, and SSIM losses, and we combine both VGG and AlexNet LPIPS losses to achieve superior visual quality.
Additionally, we empirically find that discarding the encoder of TexVAE hinders feature continuity, leading to inferior 2DGS results.

\subsection{Latent Flow Matching (FM) Diffusion Models}

\subsubsection{Structural Flow Matching (StructFM).}
\label{sec:structflow}

\begin{figure}
\centering
\includegraphics[width=1.0\linewidth]
{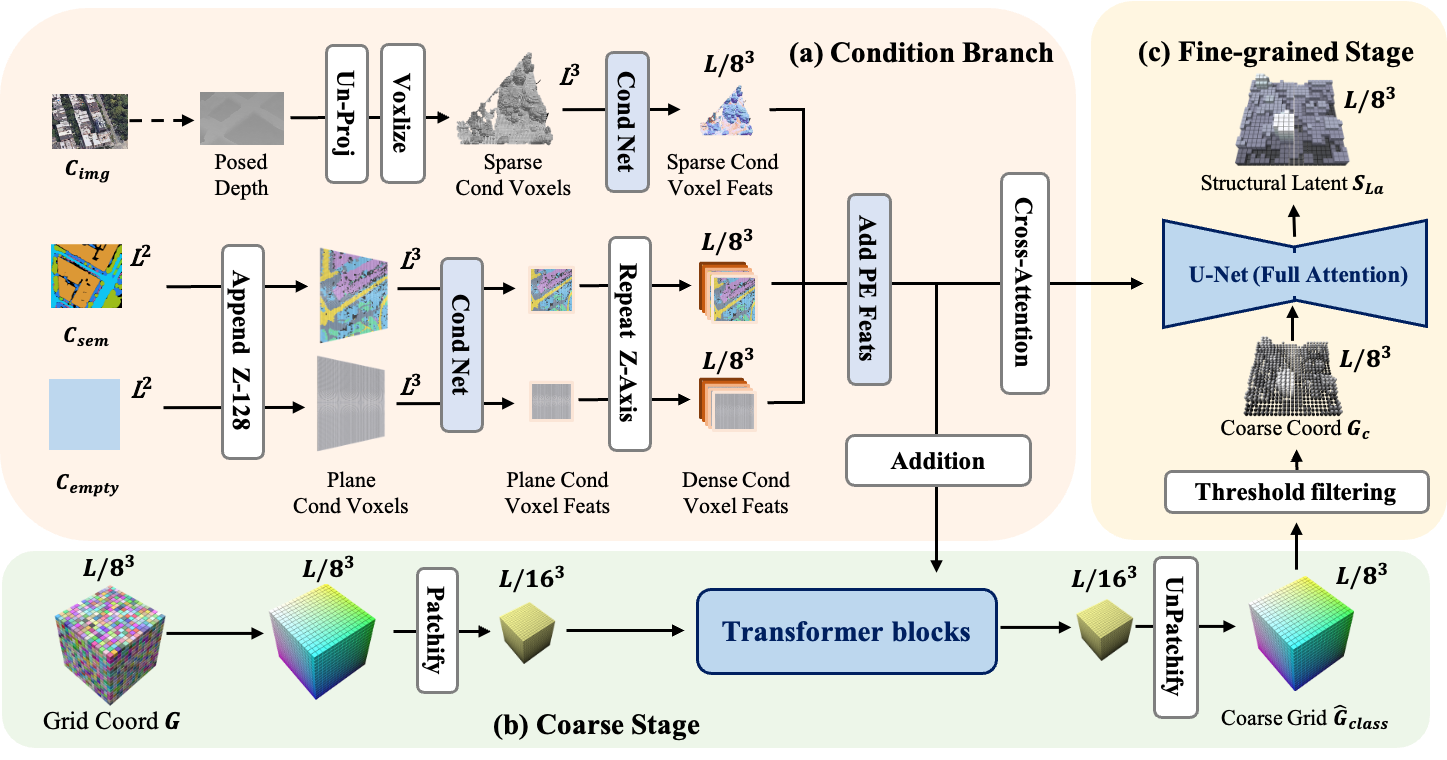}
\vspace{-0.2in}
   \caption{\textbf{Overview of the coarse-to-fine StructFM.}
   (a) Condition branch of StructFM, which receives optional inputs: image, semantic, or empty conditions.
   (b) The coarse stage is devoted to classifying activated voxels.
   (c) The fine-grained stage focuses on refining voxel coordinates and predicting structural latents based on the outcome from the coarse stage.
   \label{fig:structfm}}
\vspace{-0.25in}
\end{figure}

To handle both accurate voxel classification and structural latent feature prediction, we introduce a coarse-to-fine framework as shown in Figure~\ref{fig:structfm}.
This framework consists of two stages, each of which serves distinctly different predicting objectives.
Specifically, the coarse stage is built with pure transformer blocks, focusing on classifying activated voxels, which predicts the coarse grid $\hat{G}_{class}\in\mathbb{R}^{(\frac{L}{8})^3\times 1}$. The target for this classification can be represented as a binomial distribution $G_{class}\in\{-1, 1\}$. 
So the flow matching of coarse stage models the distribution as $p(G_{class}|G,C_{cond})$, where $G\sim\mathcal{N}(0,1)$ indicates the randomly initialized noise grid; and $C_{cond}\in\{C_{img},C_{sem},\mathrm{None}\}$, denoting optional image condition $C_{img}$, semantic segmentation $C_{sem}$, or no condition-based generation.
Note that all conditions should be projected into a 3D space. 
For this projection, monocular depth is estimated from the image condition, while semantic and non-condition inputs are assigned a dummy depth, \textit{i.e.}, z=128.
The conditional network (CondNet) is built within Swin transformer blocks~\cite{liu2021swin}, utilizing addition or cross-attention mechanisms to align the condition with voxel features.

Based on the coarse grid results $\hat{G}_{class}\in\{-1,1\}$, we set the threshold of 0, where values greater than 0 indicate valid voxels, while those less than or equal to 0 are deemed invalid.
For the fine-grained stage, our model is built within a Swin attention~\cite{liu2021swin} based U-Net, which takes the threshold filtered results $G_c$ as the sparse input, and further predicts the structural latent $S_{La}\in\mathbb{R}^{(\frac{L}{8})^{3}\times c_s}$, where $c_s=32$.
It is crucial to note that the outcomes from this fine-grained stage can also be used to refine the structural coordinates. 
We set the features of invalid voxels to zero, retaining only those voxels where more than 50\% of their channels exceed the threshold of $S_{La}>0.3$.
Additionally, to alleviate the domain gap between two stages, we propose voxel dilation augmentation to strengthen the training of the fine-grained stage.
Overall, the coarse-to-fine learning of StructFM substantially enhances structural precision, as confirmed by our experiments.

\subsubsection{Textural Flow Matching (TexFM).}
\label{sec:texflow}

Given voxel coordinates $V_c\in\mathbb{R}^{L^{3}\times 3}$ decoded from StructVAE, TexFM produces textural latent features $T_{La}\in\mathbb{R}^{L^{3}\times c_t}$ with $c_t=8$ channels.
To overcome the computational challenges associated with large-scale texture generation (up to 0.22 million voxels per scene), we enhance the efficiency of the TexFM model through a specialized U-Net architecture.
Our approach integrates sparse Swin transformer layers with full-attention layers to effectively focus on local and global feature learning, respectively.
We empirically find that Swin attention performs well in capturing high-resolution features at scales of 1/1, 1/2, and 1/4, while full attention provides a broader receptive field for low-resolution features at a scale of 1/8.
Such a complementary U-Net architecture achieves good balance between the texture quality and learning efficiency.
Moreover, TexFM also adopts flexible conditions, including images and semantic segmentations like StructFM.
More details about TexFM are presented in our supplementary.

\section{Experiments}
\paragraph{Implementation Details.}
We employ the AdamW optimizer with a learning rate of $1 \times 10^{-4}$, following a polynomial decay policy. The models are trained for 200,000 iterations using a batch size of 64 across 32 H20 GPUs. Regarding data augmentation, we implement voxel cropping and voxel flipping for 3D sparse voxels, with the corresponding camera pose also being transformed. These two basic augmentations are applied across all model training sessions.
For TexVAE, StructVAE, and TexFM, we sample training voxels with a maximum count limit of $250k$.
To improve the hole-filling capability of StructFM, we incorporate a condition voxel dropping policy based on the condition voxel normal. 
During inference, the CFG strength and sampling steps are set to 3 and 25 separately.

\paragraph{Data Preparation.}
\label{para:data_preparation}
We begin the data preparation process with a filtered set of 50k featured scene meshes.
We first extract a central mesh of size $500 \, \text{m} \times 500 \, \text{m}$ and perform a sliding crop to obtain $9$ training meshes with size of $200 \, \text{m} \times 200 \, \text{m}$. 
These training meshes are then voxelized with a voxel count of $L=360$ to generate voxel features \( V_{\text{feat}} \), where each voxel represents an area of $0.56 \, \text{m}^3$, calculated as ${200}/{360}$. This process results in $450k$ voxel features, and each voxel feature contains $220K$ voxels on average, which is $10$ times larger than Trellis voxel count.
Next, we construct a \textit{global} training and validation dataset through height sampling, yielding $447k$ training and $3,068$ validation samples. Additionally, an \textit{ablation} dataset is sampled from the New York region, consisting of $3k$ training items and $300$ validation items.


\subsection{Results of Generation}

\begin{figure}
\centering
\includegraphics[width=1.0\linewidth]
{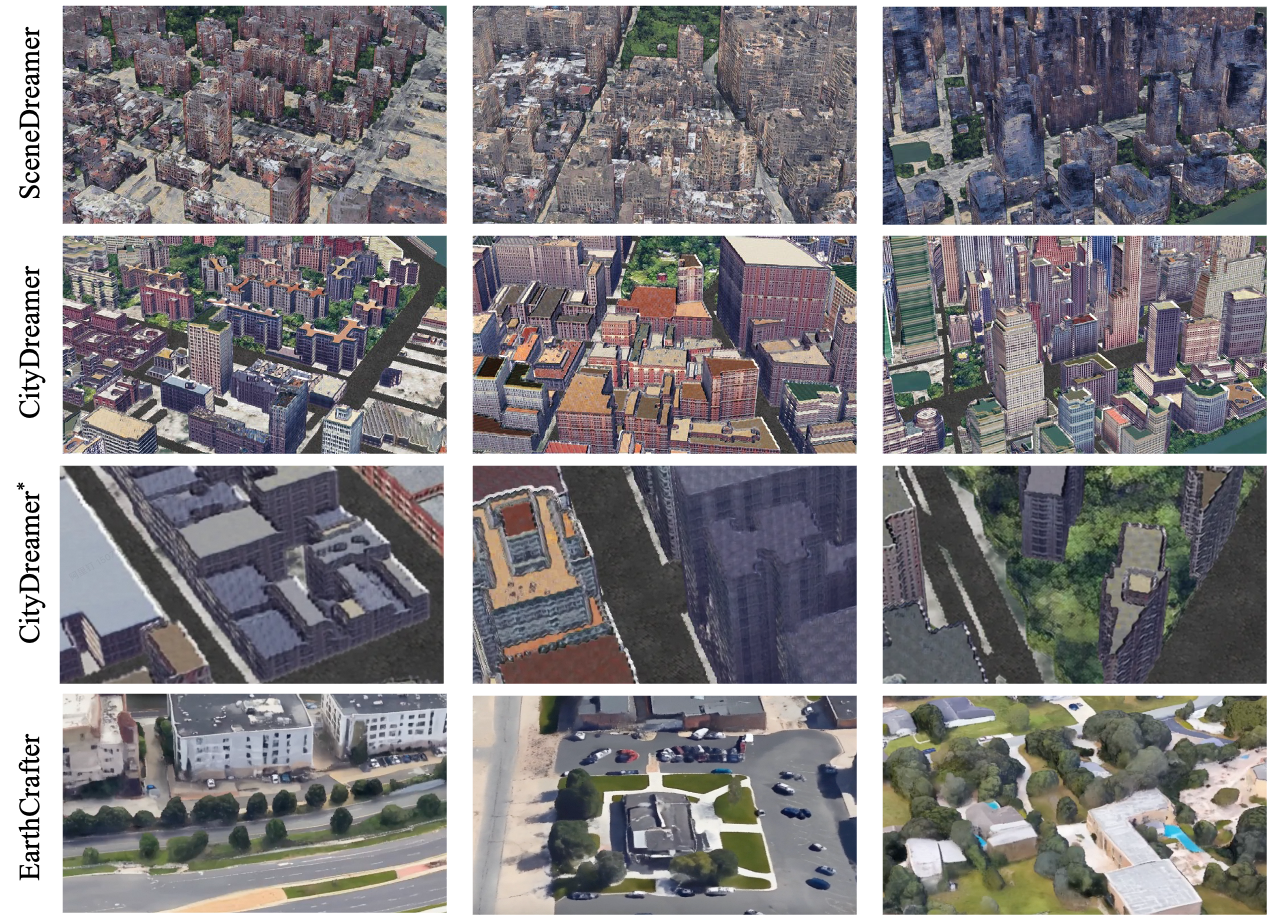}
\vspace{-0.25in}
   \caption{\textbf{Qualitative Comparison.}
   CityDreamer{*} means that showing results with similar camera distances compared to EarthCrafter.
    \label{fig:compare_others}}
\vspace{-0.2in}
\end{figure}

\paragraph{Qualitative Comparison.}
In this section, we first compare our generation results to other methods. As we focus on BEV scene generation under various conditions. To our knowledge, no research shares similar settings to ours. For example, SCube uses multiple images to generate an FPV scene, and CityDreamer uses strong semantic 3D geometry conditions, which are lifted from a 2D height map and a semantic map. So we just provide qualitative comparisons.
Figure~\ref{fig:compare_others} shows the results, and the results of EarthCrafter are generated under 2D semantic map conditions without the height condition.
Through qualitative comparison, we observe that SceneDreamer exhibits notable limitations in generating photo-realistic results, particularly manifesting in significant structural artifacts and geometric distortions in architectural elements. While CityDreamer demonstrates improved geometric fidelity at a macro level, closer inspection reveals limitations in diversity in both scene geometry and ground-level object distribution, Additionally, the textures rendered by CityDreamer tend to exhibit a somewhat cartoonish quality. In contrast, our proposed EarthCrafter framework demonstrates superior performance across multiple aspects, generating results with enhanced photo-realism and greater scene diversity compared to existing baseline methods.
Moreover, we show the qualitative results of EarthCrafter based on various conditions in Figure~\ref{fig:teaser}, demonstrating flexible capacities, and more visual results are displayed in supplementary materials.

\paragraph{Infinite Scene Generation.}
Due to the constraint of a voxel length \(L = 256\), we can generate a scene area of \(146 \, m^2\) in a single forward pass. Drawing inspiration from mask-based inpainting techniques, which leverage previously generated results to extend the scale of generation, we develop an approach to generate infinite scenes. This method utilizes a large semantic map as condition to facilitate the generation of extensive earth scenes using a sliding window manner. Constrained by GPU memory limitations in 2DGS rendering, we generate large scenes encompassing \(412 \, m^2\) with semantic size of $648 \times 648$. The results are shown in Figure~\ref{fig:large_scene_results}, and more visuals are displayed in supplementary.
\textbf{Note}: we obtain large vertical semantic map from validation patch's source scene mesh, which has overlap with training patches.
\begin{figure*}
\centering
\vspace{-0.1in}
\includegraphics[width=1.0\linewidth]
{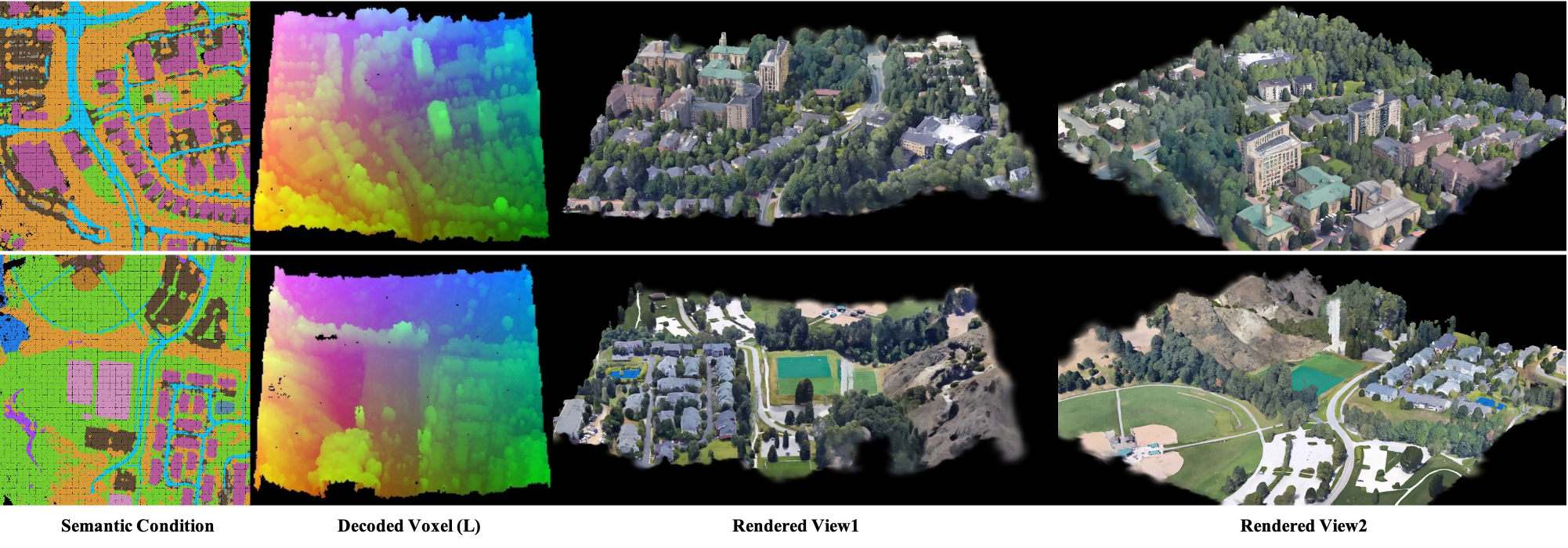}
\vspace{-0.1in}
   \caption{Infinite scene ($412m^2$) generation under large semantic condition map.
   }
\label{fig:large_scene_results}
\end{figure*}

\subsection{Ablation Studies}

\begin{table}
  \scriptsize
  \centering
  \vspace{-0.1in}
  \vspace{-0.1in}
    \begin{tabular}{c|c|c|c|c|c|cc}
    \hline 
    ID & FeatType  & L & Channel & Net & PSNR{\small{}$\uparrow$} & L1{\small{}$\downarrow$}\tabularnewline
    \hline 
     TexA & DinoSmall &  200 & 768 & Tube & 17.76 & 0.069 \tabularnewline
     TexB & $f_{0}$&        200 & 16  & Tube & 17.36 & 0.070 \tabularnewline
     TexC & $f_{0}$&        360 & 16  & Tube & 18.90 & 0.061 \tabularnewline
     TexD & $f_{0},f_{1},f_{2}$&    360 & 48  & Tube & 19.20 & 0.058 \tabularnewline
     TexE & $f_{feat}$& 360 & 66  & Tube & \textbf{19.49}  & \textbf{0.056} \tabularnewline
     \hline
     TexF & $f_{feat}$ & 360 & 66  & UNet-2 & 16.22 & 0.098 \tabularnewline
    \hline 
    \end{tabular}
  \caption{\textbf{Ablation studies of TexVAE data type and network.}\label{tab:ablation_tex_data} L denotes voxel count in training mesh, Tube represents TexVAE without any downsampling, UNET-2 means that TexVAE has two hierarchy layers.}
    \vspace{-0.1in}
\end{table}

\paragraph{TexVAE.}
We first assess voxel feature preparation policy, The experimental results are summarized in Table~\ref{tab:ablation_tex_data}, from
which we can draw the following conclusions: 
1) Comparing TexA and TexB, under the same voxel resolution $L=200$, more voxel feature channels can obtain better results.
2) Comparing TexB with TexC, we can conduct fine voxel resolution can achieve significant improvement; Comparing TexA with TexC,  we find \textit{fine-grained and low-channel voxel features are much better than features with large-channel features and coarse resolution.}
3) Comparing TexD and TexE with TexC, we find that large field features and local low-level features can continuously improve performance.
4) Results between TexE and TexF prove that voxel feature can be compressed in the channel dimension, but can't be compressed in the spatial dimension, which largely degrades performance.
Based on the above analysis, we choose mixed features $f_{feat}$ as the basic voxel feature schema, which takes 31M buffers for each $f_{feat}$ in average. In addition, because TexVAE can't be compressed in spatial, we disentangle structure and texture generation, apply a tube-shaped network for TexVAE to compress in channel dimension, and employ a U-Net-shaped network for StructVAE for spatial compression.

\begin{table}
  \footnotesize
  \centering
  \setlength{\tabcolsep}{2pt} 
    \begin{tabular}{c|c|ccc|c|c}
    \hline 
    Method & Data & PixShuffle & C-BLock  & FullAttn  & Time & Acc{\small{}{$\uparrow$}} \tabularnewline
    \hline 
      Xcube{*}& \textit{ablation} &  &  & & 29.1 & 94.3 \tabularnewline
    \hline 
      & \textit{ablation} &  &  & & 14.8 & 79.2 \tabularnewline
      & \textit{ablation} & $\checkmark$ &  & & 15.7 & 94.3 \tabularnewline
     StructVAE & \textit{ablation} & $\checkmark$ & $\checkmark$ & & 16.8 & \textbf{95.3} \tabularnewline
      & \textit{ablation} & $\checkmark$ & $\checkmark$ & $\checkmark$ & 19.5 & 94.9 \tabularnewline
      \hline
       & \textit{global} & $\checkmark$ & $\checkmark$ & & 79.3 & 96.6 \tabularnewline
      & \textit{global} & $\checkmark$ & $\checkmark$ & $\checkmark$ & 91.0 & \textbf{97.1} \tabularnewline   
    \hline
    \end{tabular}
  \caption{\textbf{Ablation results of StructVAE on ablation and train data.} \label{tab:ablation_structvae} Xcube{*} is re-implemented as StructVAE with the same blocks. PixShuffle means upsample layer in decoder, C-Block means convolution blocks, FullAttn means full attention transformer on lowest resolution layers.}
\end{table}

\paragraph{StructVAE.}
StruceVAE ablations are listed in Table~\ref{tab:ablation_structvae},
verifying the effectiveness of proposed modules. 
First, PixShuffle largely boosts the performance, which proves the importance of unambiguity upsample in sparse structure. Next, applying mixed blocks by inserting one convolution block every four Swin transformer blocks can improve performances, too many C-Blocks can't achieve continuous improvement. 
To capture global information, we utilize full attention to replace Swin attention at the lowest layer. However, it observed a decline in \textit{ablation} dataset, but shows improvement in \textit{global} training set.  We empirically think geometry in a small dataset prefers local features, but a large dataset prefers global features.
In addition, we also implement the XCube method with the same block channels, which applies a fully sparse convolution block network, as shown in Table~\ref{tab:ablation_structvae}. Our transformer based StructVAE achieves superior performance to convolution based XCube, while taking less training time.

\begin{table}
  \footnotesize
  \centering
    \begin{tabular}{c|c|c|c|c}
    \hline 
    Method & S-Num & S-Index & $mIoU^3${\small{}$\uparrow$}  & $mIoU^0${\small{}$\uparrow$} \tabularnewline
    \hline
     DenseSFM & 1 & 1 & 82.8 & 18.9 \tabularnewline
     \hline
     ClassSFM & 2 & 1 & 84.7 & - \tabularnewline
     LatentSFM & 2 & 2 & \textbf{86.1} & \textbf{25.4} \tabularnewline
    \hline 
     ClassSFM$^{\dagger}$ & 2 & 1 & 83.9 & -\tabularnewline
     LatentSFM$^{\dagger}$ & 2 & 2 & \textbf{84.3} & 23.3 \tabularnewline
    \hline
    \end{tabular}
  \caption{\textbf{Results of StructFlows on train data under image condition.} S-Num denotes total stage number, S-Index denotes stage index, $mIoU^3$ and $mIoU^0$ denote voxel structure metric at $L/8$ level and $L$ level.}
    \vspace{-0.1in}
    \label{tab:ablation_structfm}
\end{table}

\paragraph{StructFM.}


To prove the effectiveness of our two-stage sparse struct-latent generation pipeline in a coarse-to-fine manner, we implement a one-stage dense struct flow model (DenseSFM) to generate sparse struct-latents from dense noised latent volume, and DenseSFM conducts voxel classification and latent generation in one model and applies the same classification schema as the fine stage in StructFM.
Table~\ref{tab:ablation_structfm} shows the results of different structure generation methods. The pure classification flow model ClassSFM (the first stage in StructFM) achieves better accuracy than DenseSFM. This means latent generation and voxel classification have a conflict in a dense manner, which leads to performance degradation.
However, if we feed the coarse, sparse voxels from ClassSFM to LatentSFM (the second stage in StructFM), we obtain sustained performance gains in voxel classification. In total, compared to the one-stage dense manner, our two-stage method achieves significantly increased classification accuracy(+3.3).
Additionally, for the fine-level performance, our two-stage approach also obtains considerable improvement over the one-stage method. 
This proves the effectiveness of our two-stage coarse-to-fine method.

\section{Conclusion}
In this work, we introduce significant advancements in geographic-scale 3D generation through the development of Aerial-Earth3D and EarthCrafter. 
By providing the largest 3D aerial dataset to date, we have established a robust foundation for effectively modeling a diverse range of terrains and structures. 
Our dual-sparse VAE framework and innovative flow matching models not only enhance the efficiency of generating detailed textures and structures but also address key challenges associated with large-scale computation and data management.
The proposed coarse-to-fine structural flow matching model further ensures accurate structural representation while allowing for flexible conditioning based on various inputs. Additionally, we propose to use low-dimensional features from FLUX-VAE to represent voxel textures, enjoying superior reconstruction.
The rigorous experiments validate the effectiveness and superiority of our method compared to existing approaches.

\clearpage
\begin{center}
    \textbf{\huge Supplementary materials} 
\end{center}

\section{Diverse Generation}
In addition to the foundational concepts introduced in the main paper, several advanced applications merit exploration. 
To investigate the potential for diverse generation under the same conditions, three distinct experiments are outlined as follows:  

\subsubsection{Diverse Overall Generation.}
\begin{figure*}
\centering
\includegraphics[width=0.95\linewidth]
{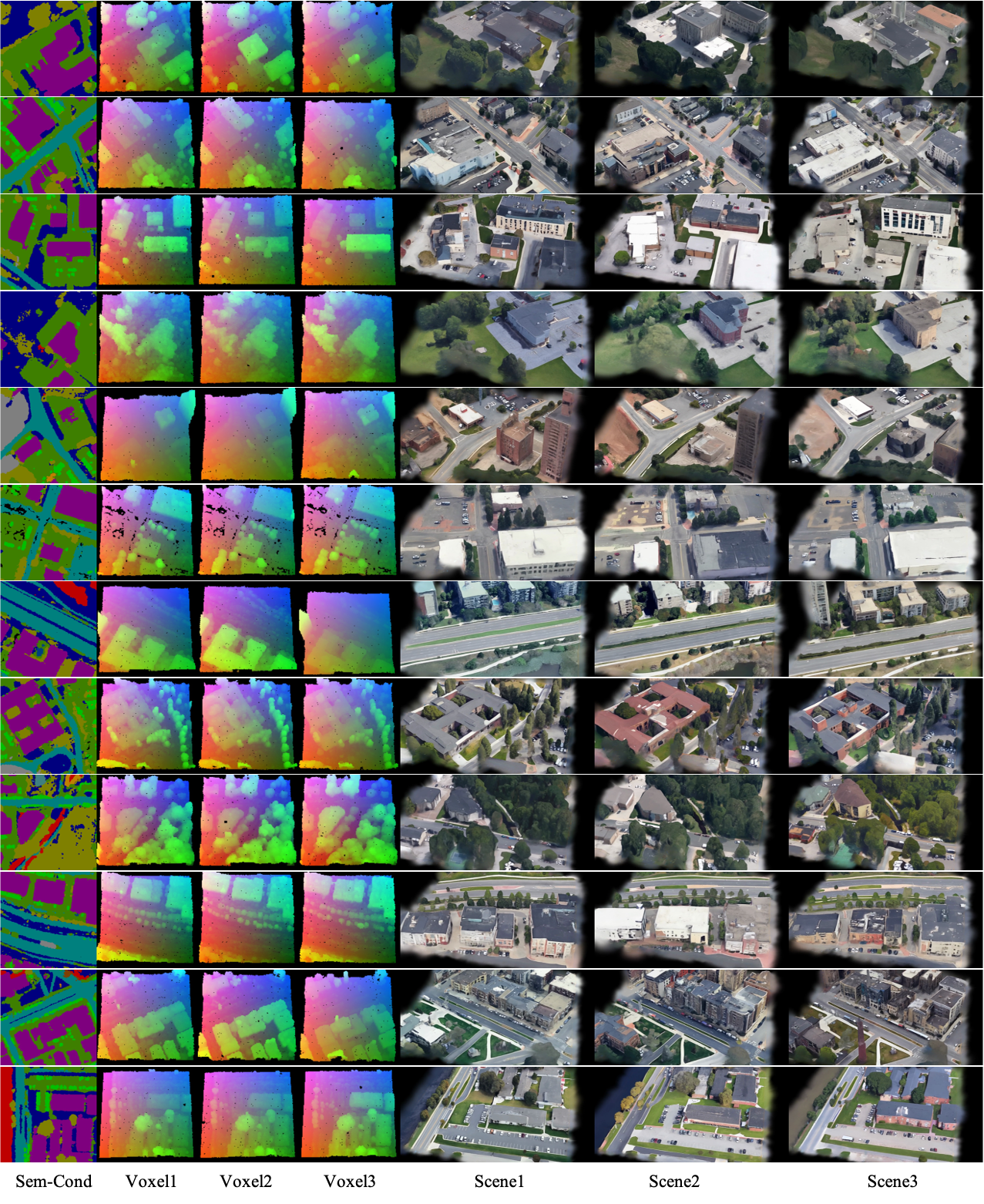}
\vspace{-0.2in}
   \caption{\textbf{Diverse scene generation under semantic condition}. \label{fig:diverse_overall_sem}}
\end{figure*}

We assess the overall capacity for diverse generation while maintaining consistent semantic conditions. The results presented in Figure~\ref{fig:diverse_overall_sem} demonstrate our ability to generate varied geometries and textures under the same semantic framework.

\subsubsection{Semantic-based Diverse Texture Generation.}
\begin{figure*}
\centering
\includegraphics[width=0.9\linewidth]
{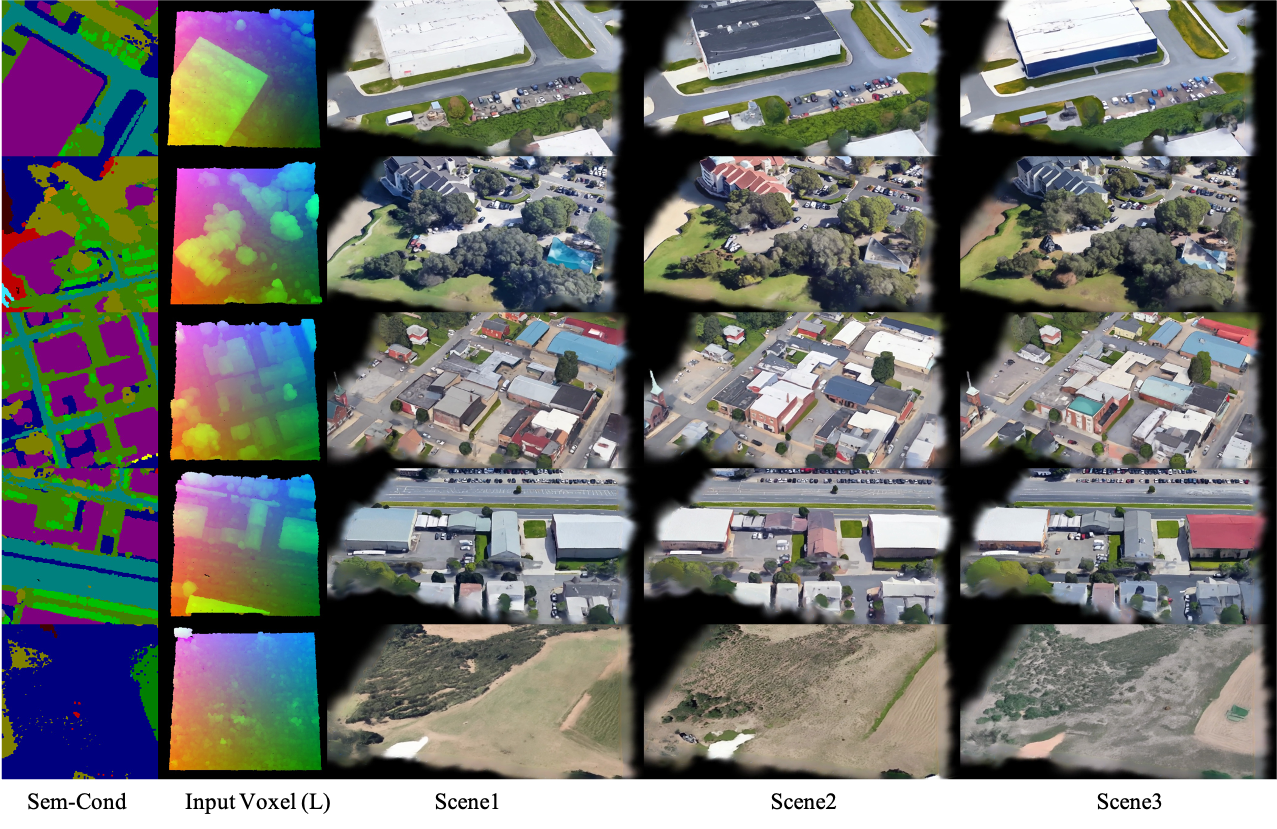}
\vspace{-0.2in}
   \caption{\textbf{Diverse texture generation under semantic condition}. \label{fig:diverse_tex_sem}}
\end{figure*}

Utilizing ground truth geometry voxels and a paired vertical-view semantic map, we generate more realistic and diverse textures. This approach facilitates the enhancement of 3D OpenStreetMap (OSM) data.  As shown in Figure~\ref{fig:diverse_tex_sem}, this method yields a variety of textures under the same geometry and semantic inputs.

\subsubsection{Unconditional Diverse Texture Generation.}
\begin{figure*}
\centering
\includegraphics[width=0.9\linewidth]
{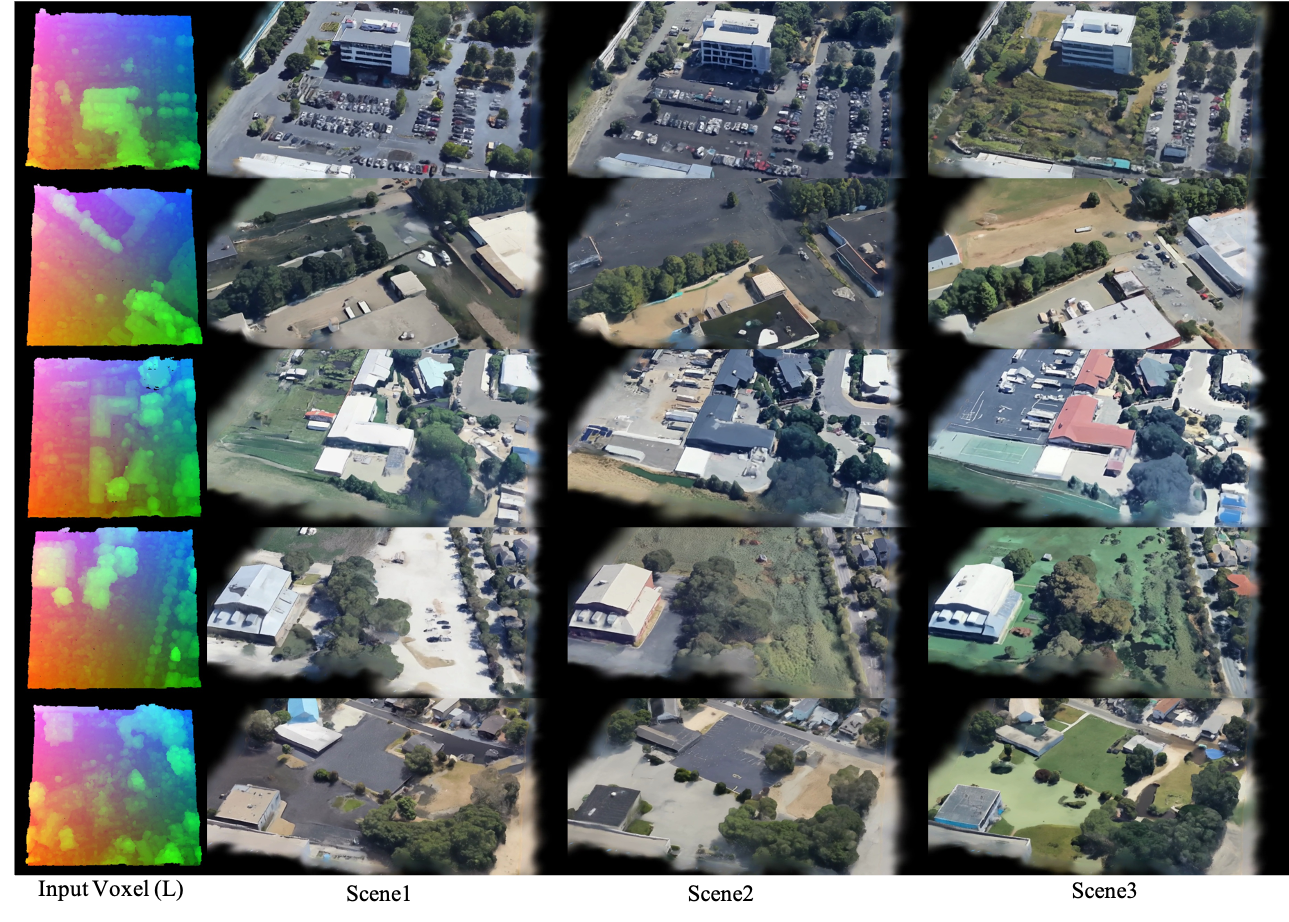}
\vspace{-0.2in}
   \caption{\textbf{Diverse texture generation without condition}. \label{fig:diverse_tex_rand}}
\vspace{-0.1in}
\end{figure*}

Using the ground truth geometry voxels without any conditions, we generate an even broader range of diverse textures. As shown in Figure~\ref{fig:diverse_tex_rand}, this method yields a variety of textures.

\subsubsection{More Generation results.}

\begin{figure*}
\centering
\includegraphics[width=0.95\linewidth]
{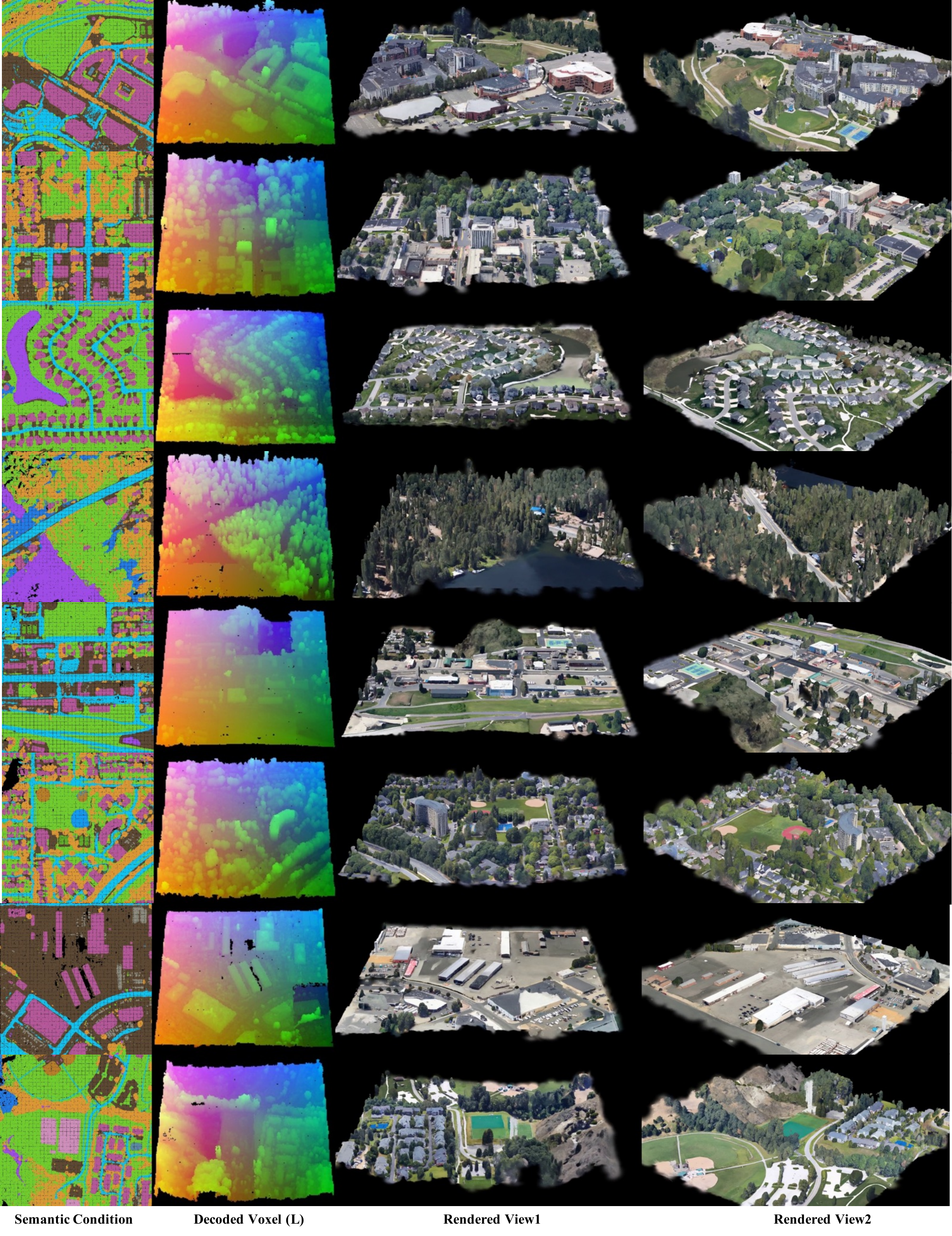}
\vspace{-0.2in}
   \caption{\textbf{Infinite scene ($412 m^2$) generation under semantic condition}. \label{fig:large_scene_results2}}
\vspace{-0.1in}
\end{figure*}

\begin{figure*}
\centering
\includegraphics[width=0.95\linewidth]
{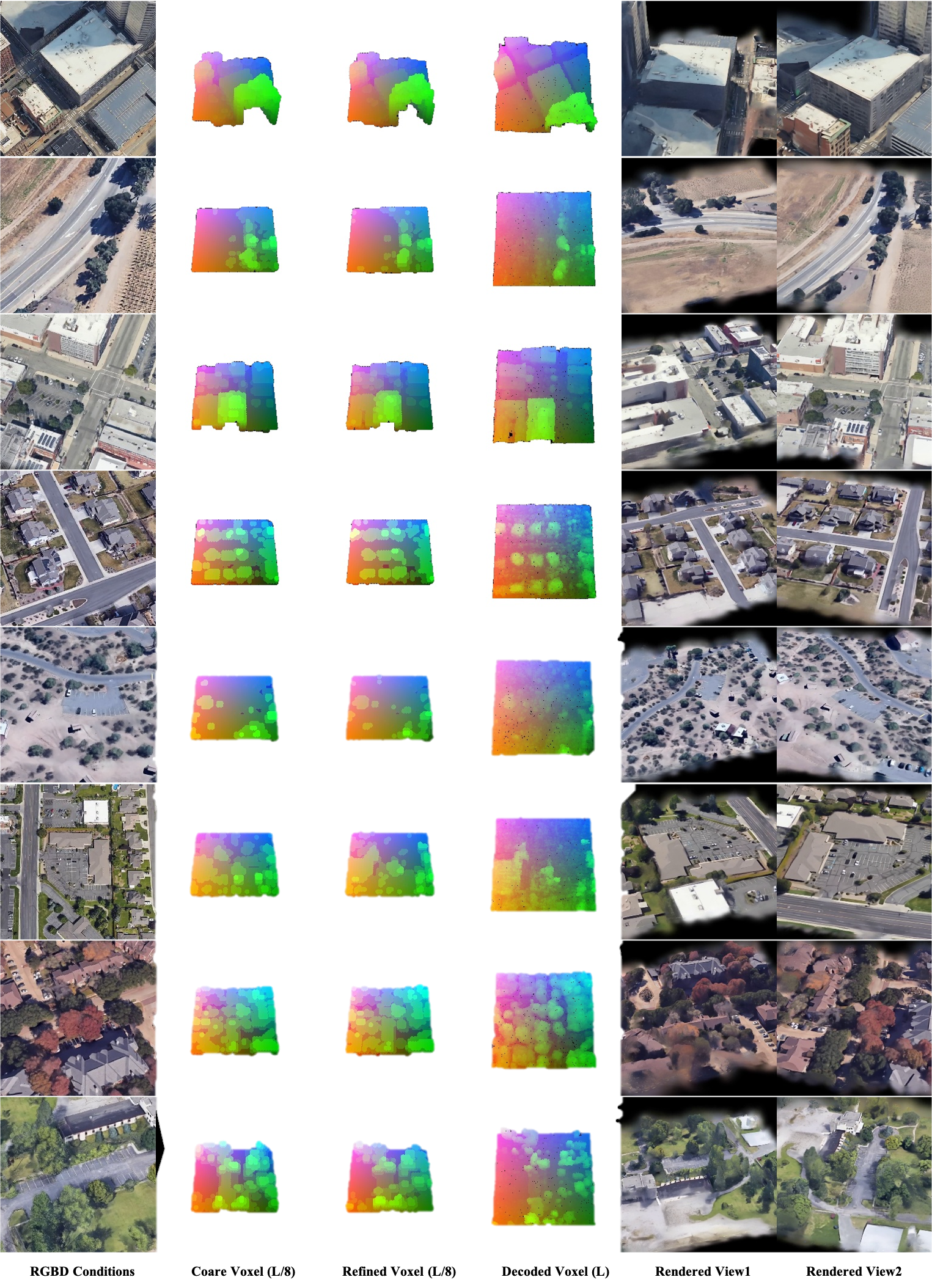}
\vspace{-0.2in}
   \caption{\textbf{Scene generation under RGBD image condition}. \label{fig:result_rgbd}}
\vspace{-0.1in}
\end{figure*}

\begin{figure*}
\centering
\includegraphics[width=0.93\linewidth]
{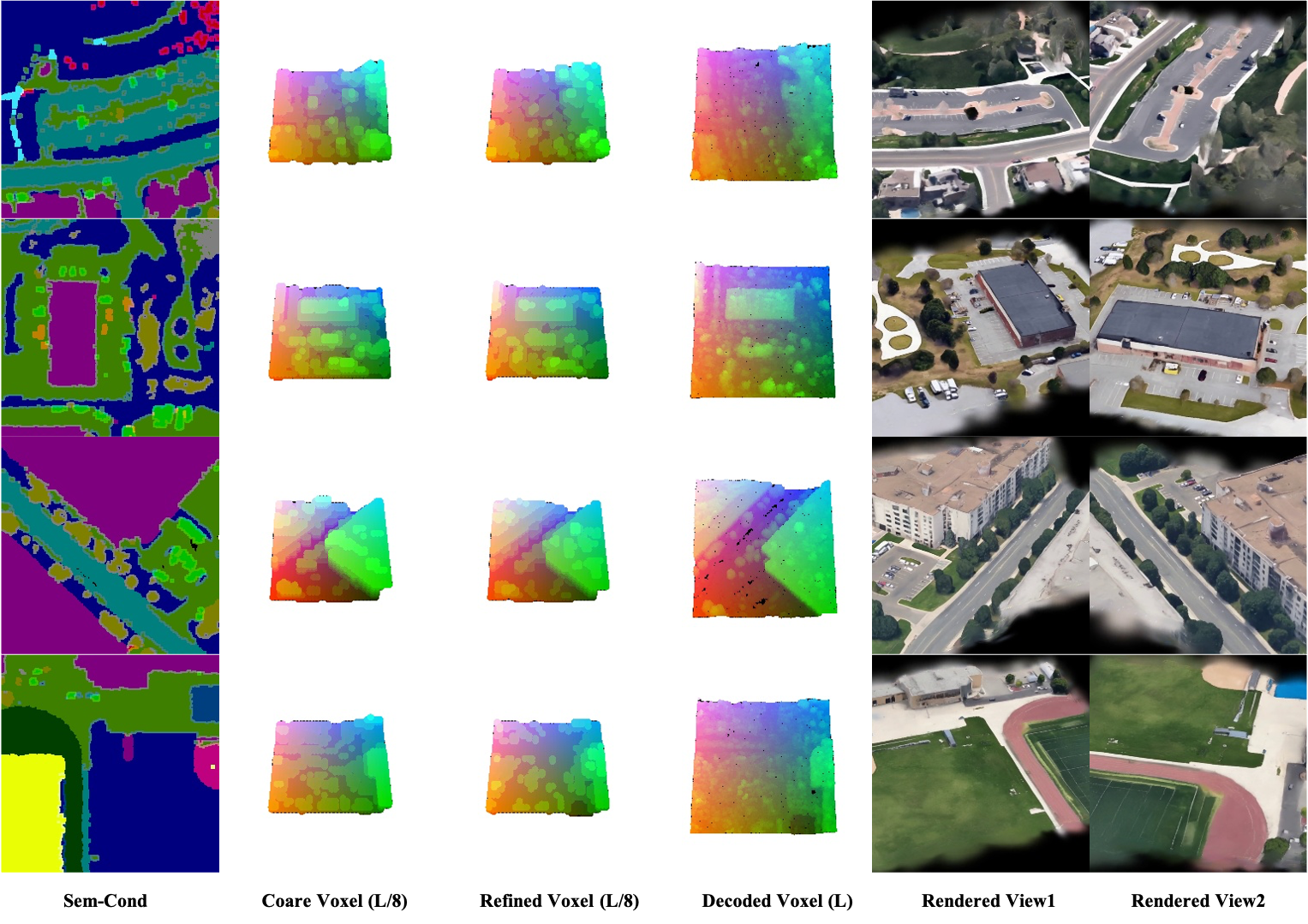}
\vspace{-0.15in}
   \caption{\textbf{Scene generation under semantic condition}. \label{fig:result_overall_sem}}
\vspace{-0.1in}
\end{figure*}

\begin{figure*}
\centering
\includegraphics[width=0.93\linewidth]
{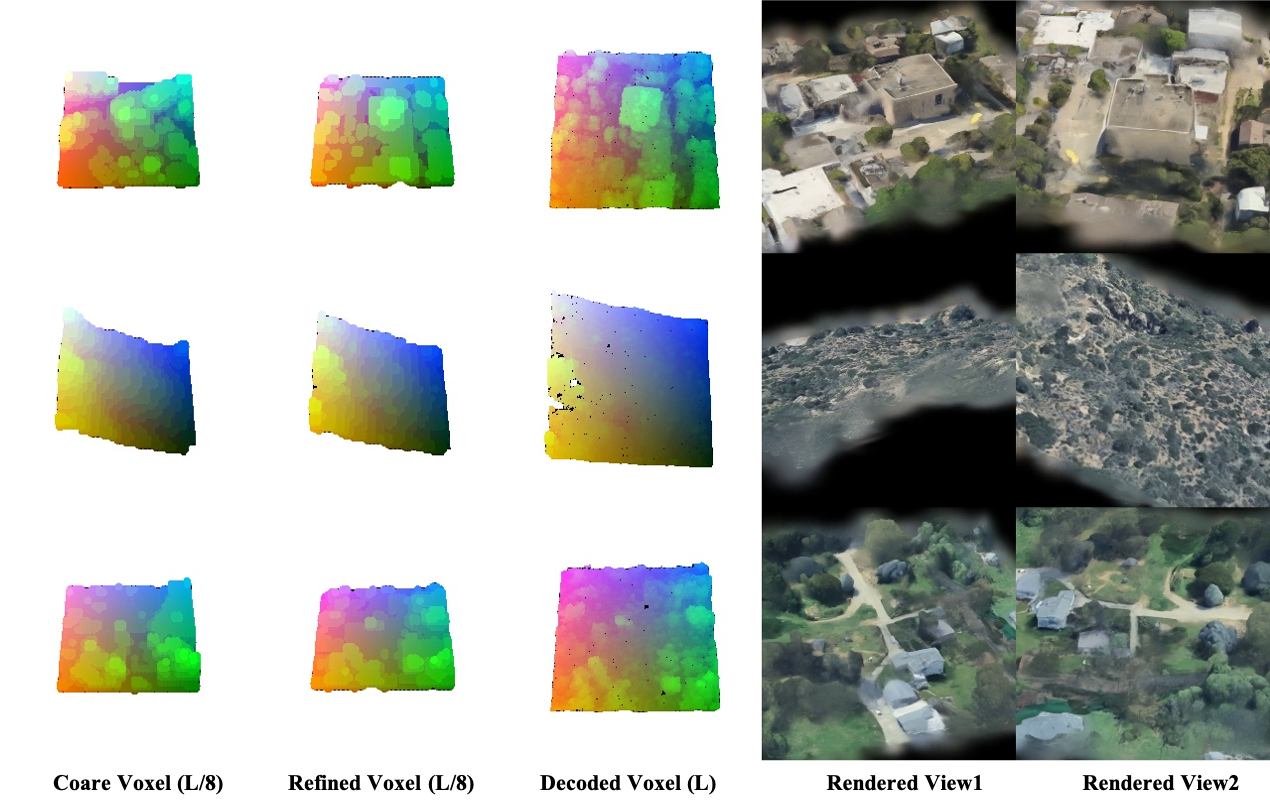}
\vspace{-0.15in}
   \caption{\textbf{Scene generation without condition}. \label{fig:result_overall_rand}}
\vspace{-0.1in}
\end{figure*}

We present additional examples of 3D scenes generated by EarthCrafter. These include more semantic condition results Figure~\ref{fig:result_overall_sem} , more RGBD condition results Figure~\ref{fig:result_rgbd}, more empty(random) condition results Figure~\ref{fig:result_overall_rand} and more infinite scene generation results Figure~\ref{fig:large_scene_results2}.


\section{Details of Model Architectures}

\subsubsection{TexVAE.}
TexVAE employs a tube-shaped encoder-decoder architecture, notably designed without any downsampling or upsampling layers. The encoder transforms the voxelized features $V_{\text{feat}} \in \mathbb{R}^{L^3 \times 66}$ into textural latent representations $T_{\text{La}} \in \mathbb{R}^{L^3 \times c_t}$, where $c_t=8$. This transformation is achieved using 12 3D-Sparse-Shift-Window Transformer (SWT-block) blocks~\cite{xiang2024structured}, each configured with 512 feature channels. Conversely, the decoder translates these latent features into 2DGS representations, sharing an identical architectural design.
In the final 2DGS representation layer, each voxel spawns 16 Gaussian primitives. Each Gaussian primitive is characterized by 23 associated parameters: its offset $o_i \in \mathbb{R}^3$, scaling $s_i \in \mathbb{R}^3$, opacity $\alpha_i \in \mathbb{R}^1$, rotation quaternion $R_i \in \mathbb{R}^4$, and spherical harmonics coefficients $c_i \in \mathbb{R}^{12}$ (corresponding to a spherical harmonic degree of 1).

\begin{table}[h]
  \small
  \centering
  \vspace{-0.15in}
  \setlength{\tabcolsep}{3.5pt} 
    \begin{tabular}{ccccc}
    \toprule 
    Model & Levels & Architecture & Params \tabularnewline
    \midrule
    TexVAE & $L/1$ & Tube & 75.8M \tabularnewline
     StructVAE & $L/1$, $L/2$ ,$L/4$ ,$L/8$  & UNet-4 & 258.5M \tabularnewline
     TexFM & $L/1$, $L/2$ ,$L/4$ ,$L/8$  & UNet-4 & 1191.7M \tabularnewline
     ClassSFM & $L/8$  & UNet-2 & 657.4M \tabularnewline
     LatentSFM & $L/8, L/16$  & UNet-2 & 1092.4M \tabularnewline
     \bottomrule 
    \end{tabular}
  \caption{Summary of Model configurations.}
    \label{tab:model_config}
  \vspace{-0.2in}
\end{table}

\subsubsection{StructVAE.}
StructVAE utilizes an encoder-decoder framework to compress the dense voxel grid coordinates $V_c \in \mathbb{R}^{L^3 \times 3}$ into structural latent representations $S_{\text{La}} \in \mathbb{R}^{(\frac{L}{8})^{3} \times c_s}$, where $c_s=32$. The encoder of StructVAE performs three downsampling operations, resulting in four hierarchical feature levels.

\begin{table}[ht]
  \small
  \centering
  \vspace{-0.1in}
  \setlength{\tabcolsep}{3.5pt} 
    \begin{tabular}{c|c|c|c|c|c}
    \toprule
    Level & T-Blocks & Attention & C-Blocks & Channels & D-Sample \\
    \midrule
    Level 0 & 2 & 3D Swin & 1 & 128 & $\checkmark$\\
    Level 1 & 2 & 3D Swin & 1 & 256 & $\checkmark$ \\
    Level 2 & 3 & 3D Swin & 1 & 512 & $\checkmark$ \\
    Level 3 & 4 & Full & 1 & 1024 & - \\
    \bottomrule
    \end{tabular}
  \caption{StructVAE endcoder configuration. T-Block: transformer block; C-Block: convolution residual block; D-Sample: voxel downsampling.}
    \label{tab:structvae_encoder}
\end{table}

Within each encoder level, the processing sequence involves four transformer blocks (T-blocks), followed by one 3D sparse convolution residual block (C-block). Subsequently, a 3D sparse convolution layer with a stride of 2 is applied for voxel downsampling, and this layer is implemented using the FVDB library ~\cite{Williams_2024}. Detailed configurations are listed in Table~\ref{tab:structvae_encoder}.

In the decoder, we introduce the novel Pseudo-Sparse to Sparse (PSS) block, designed to enable precise upsampling of sparse geometry at each decoder level. The PSS block operates by first utilizing a \emph{sparse pixel shuffle} layer to upsample the input sparse voxels, generating an initial set of pseudo-sparse voxels. Subsequently, $N^l_p$ T-blocks are applied to these pseudo-sparse voxels. A classification head, comprising two lightweight T-blocks (each with 64 channels), then predicts the activation status of these pseudo-voxels, a prediction used to prune inactive or negative voxels and thereby refine the sparse representation. Finally, $N^l_s$ T-blocks are applied to extract refined sparse voxel features. Detailed configurations of the StructVAE decoder are listed in Table~\ref{tab:structvae_decoder}.

\begin{table}[ht]
  \small
  \centering
  \setlength{\tabcolsep}{3.5pt} 
    \begin{tabular}{c|c|c|c|c|c|c}
    \toprule
    Level & $N_p$ & $N_s$ & Attention & C-Blocks & Channels & U-Sample \\
    \midrule
    Level 0 & 4 & 0 & 3D Swin & 1 & 128 & -\\
    Level 1 & 2 & 2 & 3D Swin & 1 & 256 & $\checkmark$ \\
    Level 2 & 2 & 4 & 3D Swin & 1 & 768 & $\checkmark$ \\
    Level 3 & 0 & 4 & Full & 1 & 1024 & $\checkmark$ \\
    \bottomrule
    \end{tabular}
  \caption{StructVAE decoder configuration. $N_p$ denotes the transformer blocks used in pseudo-sparse voxels, while $N_s$ refers to the transformer blocks utilized in sparse voxels. U-Sample: voxel upsampling.}
    \label{tab:structvae_decoder}
    \vspace{-0.1in}
\end{table}


\subsubsection{StructFM.}
\begin{figure}
\centering
\includegraphics[width=0.46\textwidth]{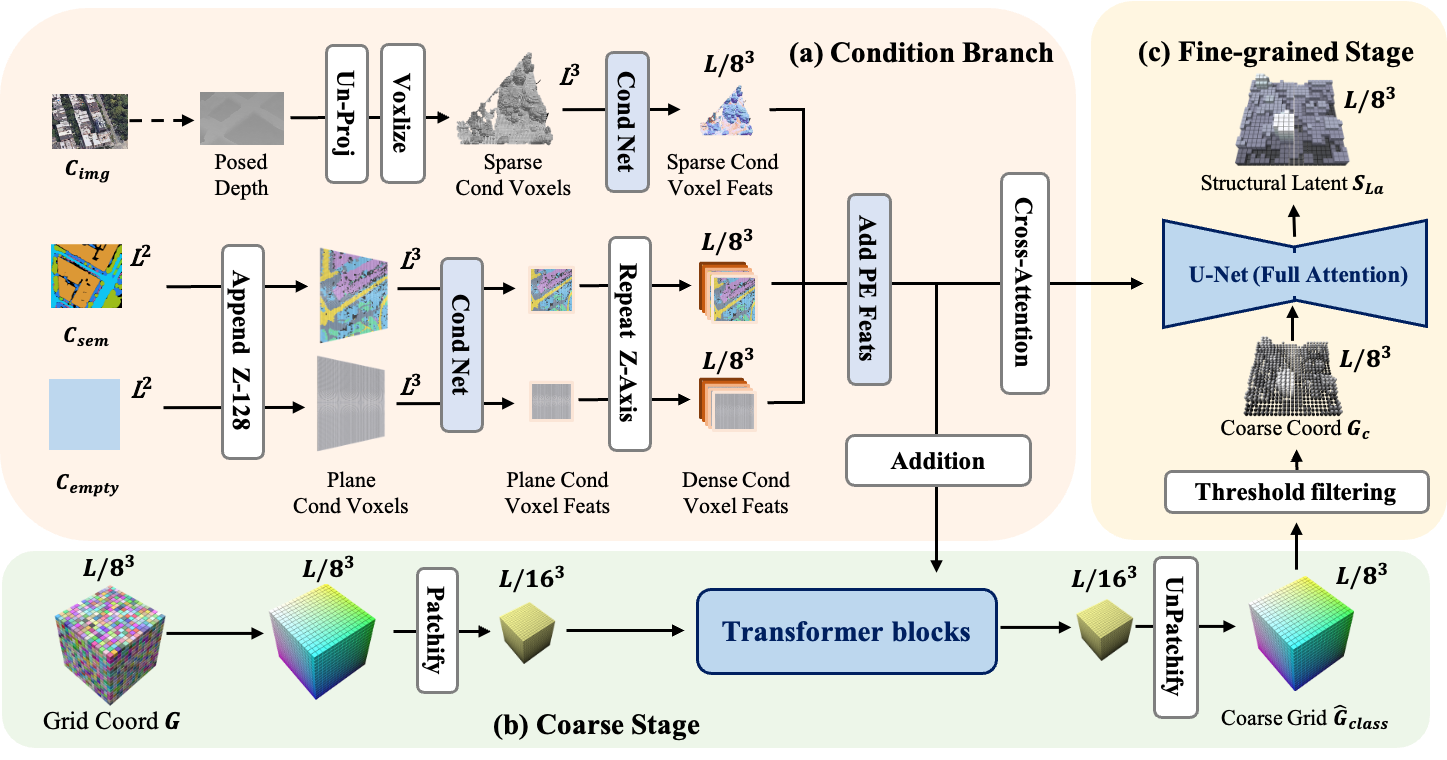}
\vspace{-0.1in}
\caption{StructFM architectures. (a) The condition branch of StructFM. (b) The coarse voxel classification stage, denoted as ClassSFM in SturctFM. (c) The voxel refinement and latent generation, called LatentSFM in StructFM.}
\label{fig:structfm_cond_net}
\end{figure}

StructFM consists of three parts: the condition branch is a multi-level encoder network, the coarse stage (ClassSFM) is a DIT-like network with pure transformer blocks, focusing on classifying activated voxels. The fine-grained stage (LatentSFM) is a U-Net-like architecture aiming to refine first-stage coarse voxels and generate voxel latent feature, which is shown in Figure~\ref{fig:structfm_cond_net}.

\emph{1) Condition Branch}
We first introduce the types of conditions and the associated condition network. There are three categories of conditions: aligned posed image condition \(C_{img}\), unaligned semantic condition \(C_{sem}\), and empty condition \(C_{empty}\).
The aligned posed image condition \(C_{img}\) can be an RGBD or RGB image; in the case of using an RGB image, the depth must be estimated using a depth prediction model, such as VGGT. The unaligned semantic condition \(C_{sem} \in \mathbb{L}^2\) is represented as a 2D semantic map from a vertical view, while the empty condition \(C_{empty}\) is also represented as a 2D vertical view semantic map with zero semantics.
It is essential to note that all conditions need to be projected into a 3D space. For the aligned condition \(C_{img}\), we project the depth information into a 3D space to generate a point cloud, which is then voxelized with a voxel length of \(0.56 \, m\) to obtain the condition voxel coordinates \(V^C_c \in \mathbb{R}^{L^3 \times 3}\). These coordinates are subsequently fed into condition network to produce the final condition voxel features \(S^C_{La} \in \mathbb{R}^{\left(\frac{L}{8}\right)^3 \times C_s}\), characterized by sparse voxels, where \(C_s\) denotes the number of feature channels, set to 768.
For unaligned conditions $C_{sem}$ and $C_{empty}$, we expand 2D semantic map a z-axis to generate 3D plane semantic condition voxels $V^S_c \in \mathbb{R}^{L^3 \times 3}$. In experiments, we set the z-axis to half of $L$, and $L$ is set to 256. After that, we feed  $V^S_c$ to the condition network, which produces plane condition voxel features, then repeat the plane voxel feature $L / 8$ times to form condition voxel features $S^C_{La}$ with dense voxels. 
Subsequently, we augment \(S^C_{La}\) with additional voxel position features, which are computed by passing the position embeddings of the condition voxel coordinates of $S^C_{La}$ through a linear layer. Finally, the condition voxel features are injected into the ClassSFM and LatentSFM networks through addition or cross-attention mechanism.

The condition network comprises four hierarchical levels, each integrating transformer blocks. Downsampling between levels is achieved through average pooling with a stride of 2. The specific configurations of the condition network are listed in Table~\ref{tab:StructFM_cond}.
\begin{table}[t]
  \small
  \centering
  \vspace{-0.1in}
  \setlength{\tabcolsep}{3.5pt} 
    \begin{tabular}{c|c|c|c|c}
        \toprule
        Level & Blocks & Attention & Channels & DownSample \\
        \midrule
        Level 0 & 2 & 3D Swin & 128 & $\checkmark$\\
        Level 1 & 2 & 3D Swin & 256 & $\checkmark$ \\
        Level 2 & 3 & 3D Swin & 512 & $\checkmark$ \\
        Level 3 & 4 & Full & 768 & - \\
        \bottomrule
    \end{tabular}
  \caption{StructFM condition network configuration}
    \label{tab:StructFM_cond}
\end{table}

\emph{2) ClassSFM} ClassSFM receives randomly initialized dense voxel grids \(G \sim \mathcal{N}(0,1) \in (L / 8)^3\) as input. It first implements four transformer blocks with full attention, followed by a patchify operation on the dense features with a stride of 2. Subsequently, 24 additional transformer blocks with full attention are applied. Finally, an up-patchify operation is performed to recover the original voxel resolution, resulting in the final prediction \(\hat{G}_{class} \in \{-1, 1\}\). Specifically, all transformer blocks are configured with 1024 channels, and after patchifying, cross-attention condition injection is conducted every 8 transformer blocks.

\emph{3) LatentSFM}
LatentSFM takes the predicted coarse voxels \(\hat{G}_{c} \in (L / 8)^3\) as inputs. It first employs a C-block and a T-block with full attention as pre-blocks, followed by the insertion of a T-block with cross attention to integrate condition features. Subsequently, a two-level hierarchical U-Net with skip connection processes the data further. Each U-Net level starts with a C-block, followed by several T-blocks with full attention, and concludes with another C-block. Moreover, at the feature merge layer between the first and second levels of the UNet, a second cross-attention T-block is introduced to further integrate condition features.
LatentSFM performs two simultaneous tasks: coarse voxel refinement and structural voxel latent generation. 
More detailed configurations are detailed in Table~\ref{tab:LatentSFM_net}.

\begin{table}[t!]
  \small
  \centering
  \vspace{-0.1in}
  \setlength{\tabcolsep}{3.5pt} 
    \begin{tabular}{c|c|c|c|c}
    \toprule
    Level & T-Blocks & Attention & C-Blocks & Channels  \\
    \midrule
    Level 0 & 4 & Full & 2 & 512 \\
    Level 1 & 22 & Full & 2 & 1280 \\
    \bottomrule
    \end{tabular}
    \vspace{-0.1in}
  \caption{LatentSFM network configuration. T-Block: transformer block; C-Block: convolution residual block.}
    \label{tab:LatentSFM_net}
\end{table}

\subsubsection{TexFM.}
\begin{figure}
\centering
\includegraphics[width=0.46\textwidth]{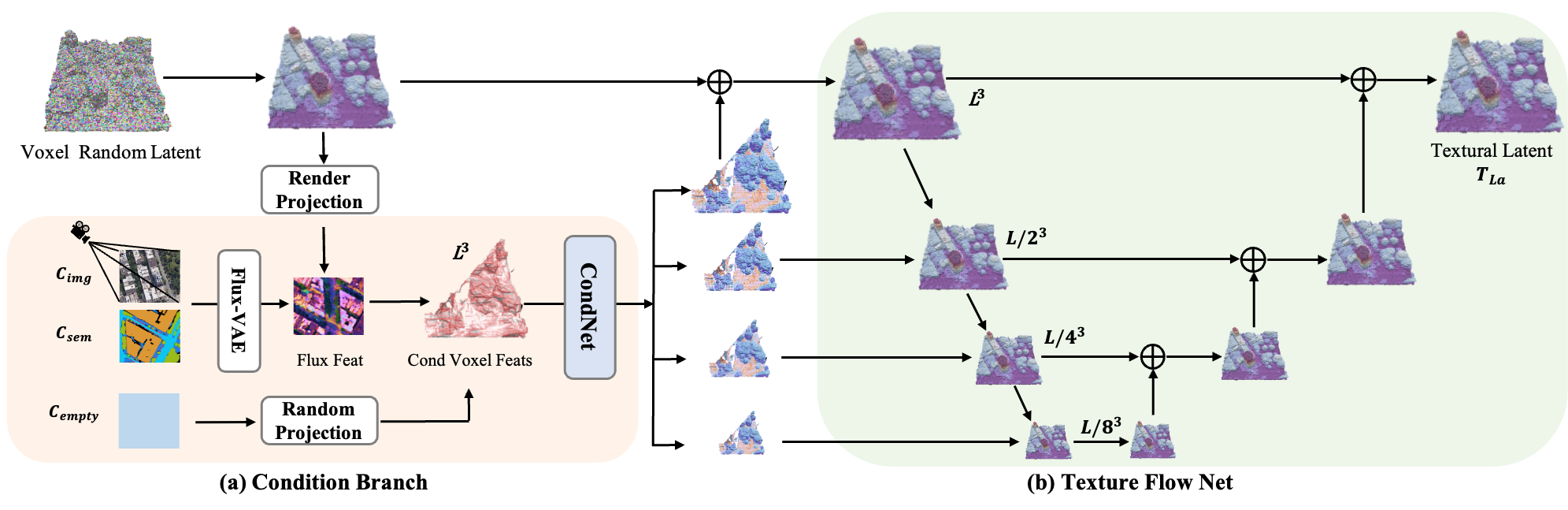}
\vspace{-0.1in}
\caption{TexFM architectures. (a) The condition branch of TexFM. (b) The main network of TexFM.}
\vspace{-0.1in}
\label{fig:texfm_net}
\end{figure}

TexFM adopts flexible conditions to produce textural latent features $T_{La}\in\mathbb{R}^{L^{3}\times c_t}$ with $c_t=8$ channels. TexFM consists of two parts: the condition network (CondNet) and the texture flow network (TexNet).
The condition branch is a multi-level encoder network, and TexNet is a UNet-shaped network, as shown in Figure~\ref{fig:texfm_net}.

\emph{1) TexFM CondNet.}
Unlike the condition network of StructFM, the TexFM condition utilizes rendering projection to map 2D Flux features \(C_f\) into 3D space, where \(C_f\) is produced by the Flux VAE encoder. In our experiments, we first convert the 2D semantic map \(C_{sem}\) into a semantic RGB image \(C_{srgb}\) using a semantic ID color mapping dictionary. We then feed either \(C_{srgb}\) or \(C_{img}\) into CondNet to obtain \(C_f\).
Specifically, given the camera pose of \(C_{img}\), we utilize a pinhole rendering model to render \(V_c \in \mathbb{R}^{L^3}\) and establish the projection relationship between the image features \(C_f\) and \(V_c\), thereby yielding the initial condition voxel features \(T_{c} \in \mathbb{R}^{L^{3} \times f_c}\), where \(f_c\) denotes the number of input condition feature channels. For the Flux features \(C_f\) derived from the vertical view \(C_{sem}\), we employ parallel up-to-down rays to build the projection relationship, realized through the voxel rendering function in the FVDB library. In the case of the empty condition, we randomly select \(10,000\) voxels from \(V_c\) and assign zero values to their features, resulting in empty condition voxel features \(T_{c}\).

Once the condition voxel features \(T_{c}\) are obtained, we feed them into CondNet to generate condition voxel features at each hierarchical level.
Similar to the condition network in StructFM, the CondNet in TexFM also comprises four hierarchical levels, each integrating transformer blocks. The specific configurations of the condition network are detailed in Table~\ref{tab:TexFM_cond}.

\begin{table}[t!]
  \small
  \centering
  \vspace{-0.1in}
  \setlength{\tabcolsep}{3.5pt} 
    \begin{tabular}{c|c|c|c}
        \toprule
        Level & Blocks & Attention & Channels \\
        \midrule
        Level 0 & 2 & 3D Swin & 192  \\
        Level 1 & 2 & 3D Swin & 384 \\
        Level 2 & 3 & 3D Swin & 768 \\
        Level 3 & 5 & Full & 1280 \\
        \bottomrule
    \end{tabular}
    \vspace{-.1in}
    \caption{TexFM condition network configuration}
    \label{tab:TexFM_cond}
\end{table}

\emph{2) TexNet.}
The network of TexNet is similar to LatentSFM; TexNet first employs a C-block and a T-block with 3D swin attention as pre-blocks. The main architecture is a four-level hierarchical U-Net with skip connections. Each U-Net level starts with a C-block, followed by several T-blocks, and concludes with another C-block. 
As TexNet takes $0.22$ million voxels as inputs, which is significantly larger than the input voxel number of LatentSFM.
In contrast to LatentSFM, which uses cross-attention to integrate condition features. TexNet applies a simple addition to merge conditional features, and we design an efficient voxel coordinates alignment algorithm between part condition voxels and global scene voxels to support efficient condition adding and merging.
The configurations of TexNet are detailed in Table~\ref{tab:texfm_net}.

\begin{table}[t!]
  \small
  \centering
  \vspace{-0.1in}
  \setlength{\tabcolsep}{3.5pt} 
    \begin{tabular}{c|c|c|c|c|c}
    \toprule
    Level & T-Blocks & Attention & C-Blocks & Channels & D-Sample \\
    \midrule
    Level 0 & 4 & 3D Swin & 2 & 192 & -\\
    Level 1 & 4 & 3D Swin & 2 & 384 & $\checkmark$ \\
    Level 2 & 8 & 3D Swin & 2 & 768 & $\checkmark$\\
    Level 3 & 12 & Full & 2 & 1280 & $\checkmark$ \\
    \bottomrule
    \end{tabular}
    \vspace{-0.1in}
  \caption{TexNet network configuration}
    \label{tab:texfm_net}
\end{table}

\emph{3) TexFM Ablations.}
To enhance efficiency, we conduct detailed ablation studies on the design of TexFM. The experimental results are summarized in Table~\ref{tab:ablation_texfmd}. First, applying C-Blocks at both ends of each UNet boosts performance. Second, while adding deeper layers marginally improves performance, it also considerably increases training time. Third, enlarging the feature channels of each block leads to substantial improvements within a reasonable computational budget. Additionally, we explored condition injection using a cross-attention mechanism; however, we found minimal correspondence between the condition image and the generated result. Consequently, we opted to replace the cross-attention approach used in StructFM with an additive condition integration method.

\begin{table}
  \scriptsize
  \centering
  \vspace{-0.1in}
  \setlength{\tabcolsep}{3.0pt} 
    \begin{tabular}{cccc|c|c|c}
    \hline 
    C-BLock & Layers & Channels & Cross  & Train Time (h)  & $PSNR_{img}${\small{}{$\uparrow$}} & $FID_{sem}${\small{}{$\downarrow$}}  \tabularnewline
    \hline
      &  &  & & 48 & 17.2 & 19.9 \tabularnewline
    $\checkmark$ &  &  & & 51 & 17.3 & 19.3 \tabularnewline
     $\checkmark$ & $\checkmark$ & & & 63 & 17.2 & 18.9 \tabularnewline
    $\checkmark$ &  & $\checkmark$ & & 54 & \textbf{17.5} & \textbf{18.6} \tabularnewline
     $\checkmark$ &  & $\checkmark$ & $\checkmark$ & 53 & 14.2 & 19.7 \tabularnewline
    \hline
    \end{tabular}
  \caption{Ablation results of TexFM. C-Block: incorporating sparse convolution blocks at both ends of each U-Net level. ``Layers'' indicates more transformer blocks at each level, while ``Channels'' denotes the enlargement of block channels. ``Cross'' signifies condition integration using a cross-attention mechanism. $PSNR_{img}$ and $FID_{sem}$ are used to evaluate the performance under image-based and semantic conditions, respectively.}
    \vspace{-0.1in}
    \label{tab:ablation_texfmd}
\end{table}

\section{Training Details}
\subsection{Training Tools}
We employ the FVDB ~\cite{Williams_2024} and SparseTensor~\cite{spconv2022} libraries to perform sparse layer operations, and utilize PyTorch3D~\cite{ravi2020pytorch3d} to build the rendering projection relation between 3D sparse voxels and 2D image conditions, while utilizing GSplat~\cite{ye2025gsplat} as our 2DGS~\cite{Huang2DGS2024} rendering tool. 

\subsection{Training Loss}
\subsubsection{TexVAE Loss.}
We employ a hybrid loss to train TexVAE, which consists of L1, LPIPS~\cite{zhang2018unreasonable}, and SSIM losses. We combine both VGG and AlexNet LPIPS losses to achieve superior visual quality, which are defined by the following equation:
\[
\text{L}_{\text{texvae}} = \lambda_{\text{l1}} \cdot L_{\text{l1}} + \lambda_{\text{ssim}} \cdot L_{\text{ssim}} + \lambda_{\text{vgg}} \cdot L_{\text{vgg}} + \lambda_{\text{alex}} \cdot L_{\text{alex}},
\]
where $L_{\text{vgg}}$ denotes LPIPS loss with VGG model,  $L_{\text{alex}}$ represent LPIPS loss with AlexNet model. The weights of each loss are set to $\lambda_{l1}=20, \lambda_{ssim}=2, \lambda_{vgg}=1.4, \lambda_{alex}=0.6$ separately.

\begin{figure*}[ht!]
\centering
\includegraphics[width=0.95\linewidth]
{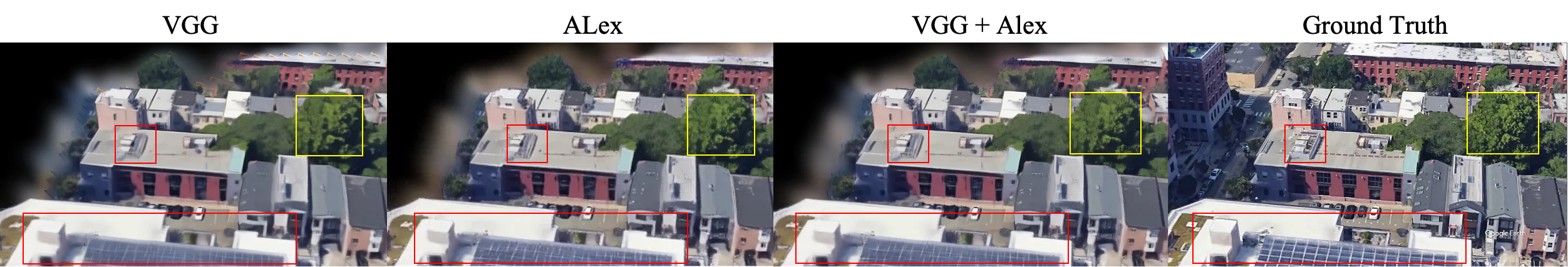}
\caption{Results of texture reconstruction changing caused by mixed LPIPS-VGG and LPIPS-ALex loss.
 VGG means using pure LPIPS-VGG loss; Alex means using pure LPIPS-Alex loss; VGG+ALex means using mixed LPIPS loss.}
 \label{fig:mixed_lpips_change}
     \vspace{-0.1in}
\end{figure*}

To evaluate the quality of texture reconstruction using different LPIPS losses, Figure~\ref{fig:mixed_lpips_change} presents the rendering results of 2DGS at the final 3D representation layer of TexVAE. Our findings indicate that TexVAE trained with the VGG LPIPS loss produces a lightly blurred texture within the red bounding box, while the tree texture in the yellow bounding box performs well. In contrast, TexVAE trained with the Alex LPIPS loss yields sharper and clearer textures in the red bounding box, but it contains white dot-shaped noise in the yellow bounding box (which becomes more apparent when zoomed out). By combining these two types of LPIPS loss, we are able to achieve improved texture quality in both the red and yellow areas.

\subsubsection{StructVAE Loss.}
The StructVAE architecture initially compresses the full voxel coordinates $V_c \in \mathbb{R}^{L^3 \times 3}$ into a latent representation $S_{La} \in \mathbb{R}^{(\frac{L}{8})^3 \times c_s}$ via three successive down-sampling operations. Subsequently, these latent features are up-sampled three times, recovering voxel coordinates $V_c \in \mathbb{R}^{L^3 \times 3}$ through our proposed Pseudo-Sparse to Sparse (PSS) block.
Within each PSS block, a voxel classification approach is employed to extract valid sparse voxels from the initially up-sampled pseudo-sparse voxels. At each voxel level $l \in \{0, 1, 2\}$, the corresponding down-sampled voxels $V^l_c$ serve as the ground truth for sparse voxels, while $V^l_p$ represents the up-sampled pseudo-sparse voxels.

During training, pseudo-sparse voxels $V^l_p$ are matched against the ground truth voxels $V^l_c$. If a pseudo-voxel $v^l_p \in V^l_p$ is also present in $V^l_c$, its corresponding classification label is set to 1, indicating an active state and this pseudo voxel should be kept. Otherwise, the label is set to 0, signifying that $v^l_p$ should be filtered.  A cross-entropy loss, $L^l_{\text{ce}}$, is used to supervise the voxel classification head at each level. The overall StructVAE loss is then defined as:
\[
L_{\text{structvae}} = \lambda_{0} \cdot L^0_{\text{ce}} + \lambda_{1} \cdot L^1_{\text{ce}} + \lambda_{2} \cdot L^2_{\text{ce}}.
\]
In our experiments, the loss weights $\lambda_l$ for each level were uniformly set to 5.  We observed that voxel levels 2 and 1 converged rapidly during training, while convergence at level 0 was delayed until the higher levels had stabilized. During evaluation, the accuracies at levels 2 and 1 reached near-perfect performance (approaching 100\%), and level 0 achieved a state-of-the-art accuracy of $97.1\%$, demonstrating the efficacy of our novel network architecture.

\subsubsection{Flow Model Loss.}
We employ a DiscreteScheduler noise schedule with scheduler shift $3.0$, and apply the same conditional flow matching objective as~\cite{xiang2024structured} to train TexFM and StructFM.

\section{Data Preparation}

\begin{figure}[t!]
\centering
\includegraphics[width=1.0\linewidth]
{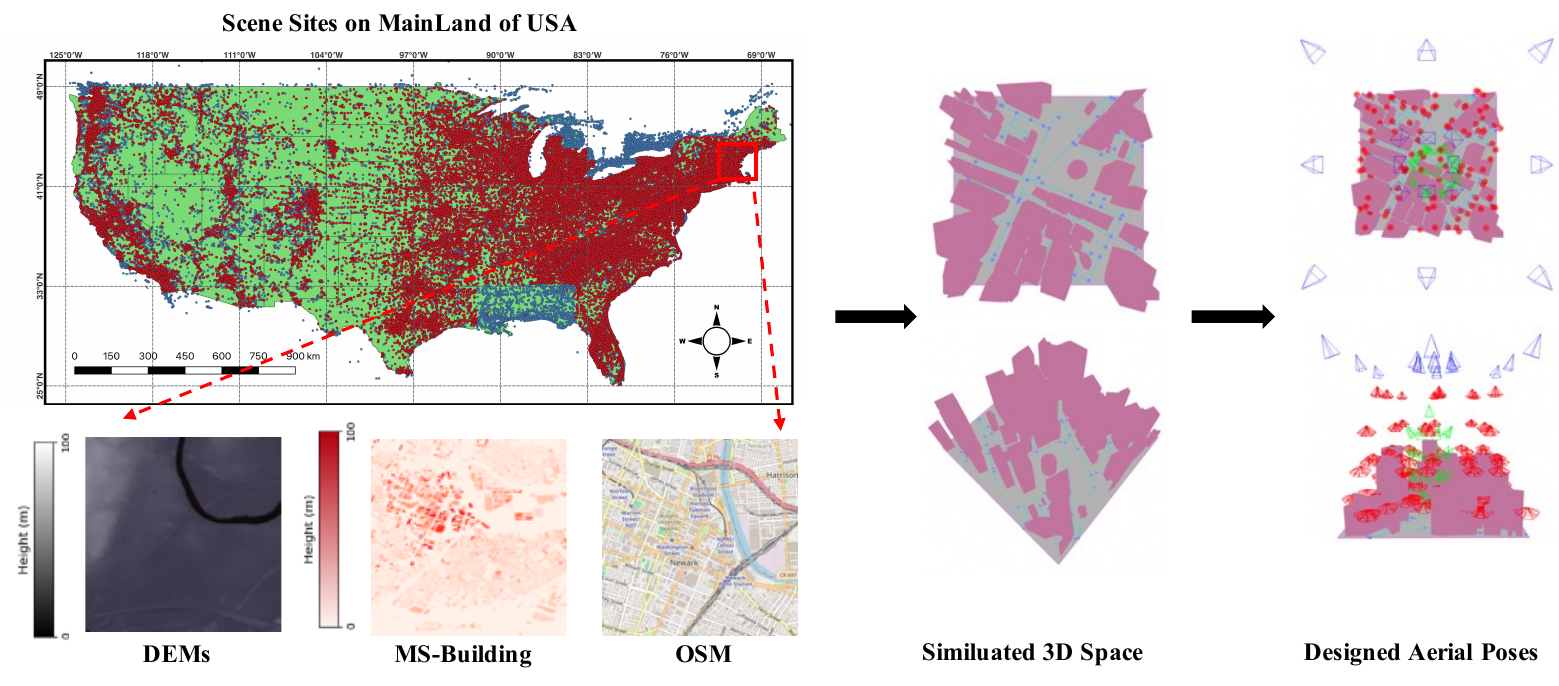}
\vspace{-0.1in}
   \caption{The illustration of data collection of Google Earth. \textbf{\textcolor{darkblue}{Blue points}} denote all 150k scenes originally captured from Google Earth, while \textbf{\textcolor{darkred}{red points}} denote the final 50k training scenes filtered for the EarthCrafter training. We employ DEM terrain~\cite{dem}, MS-Building heights~\cite{GlobalMLBuildingFootprints}, and driving road paths from OSM~\cite{openstreetmap} to build the 3D simulation for defining reasonable camera viewpoints.
    }
\label{fig:data_collection}
\vspace{-0.2in}
\end{figure}

\paragraph{Collection and 3D Simulation.}
To create a large-scale aerial dataset with representative scenes, we utilized the rating system of the Google Earth platform for scene selection. By setting a rating threshold of greater than 3.5 and employing spatial deduplication within 600m grids, we ultimately identified 150,745 high-quality points of interest across the mainland U.S., as shown in Figure~\ref{fig:data_collection}. 
Furthermore, we integrated the OSM driving roads~\cite{openstreetmap}, DEM terrain~\cite{dem}, and MS-Building~\cite{GlobalMLBuildingFootprints} height data to construct a high-precision simulated 3D scene for the following camera definition.

\paragraph{Camera Viewpoints.}
To achieve maximum scene coverage with minimal sampled perspectives while avoiding outliers such as occlusions and through-modeling, we have developed a comprehensive viewpoint planning scheme based on the simulated 3D scene. 
Specifically, we planned three complementary camera trajectory patterns centered around the latitude and longitude of each point of interest within this simulated 3D scene.
1) Firstly, a dual-layer top-down trajectory is established as \textit{Top-Pose} at a height of 500m above ground. This trajectory consists of a larger outer square measuring 600m$\times$600m and a smaller inner square of 100m$\times$100m.
Eight viewpoints are evenly assigned along the edges of each square, with cameras in the outer square oriented towards the center of the scene and those in the inner square directed toward the scene's edges, which balances capturing the overall scene layout with local details.
2) Secondly, we implemented a multi-layer spiral trajectory called \textit{Building-Pose}, centered around the tallest building in the scene, using a top-down sampling method to ensure holistic coverage of the core structure. 
3) Finally, we designed an adaptive trajectory system called \textit{AdaLevel-Pose}. 
This system creates progressively smaller squares at increasing heights, beginning at a minimum altitude of 75m and extending to a height of the tallest building plus 225m, while six surrounding viewpoints are assigned to each square.  
Additionally, an intelligent obstacle avoidance mechanism based on the driving road paths~\cite{openstreetmap} is incorporated to ensure the accessibility of sampled viewpoints.
We finally achieved 97,165 scenes from 150k with valid viewpoint assignments. 

\subsection{Achievement of Annotated Mesh}
\label{sec:mesh_achievement}



\paragraph{InstantNGP Training.}
We choose an efficient volume rendering approach based NeRF~\cite{mildenhall2021nerf}, called InstantNGP~\cite{mueller2022instant}, as the annotation way to achieve a 3D mesh for each aerial scene.
Our process begins by converting the latitude and longitude coordinate system into the widely adopted OpenGL coordinate system.
First, the latitude and longitude coordinates are converted to Earth-Centered, Earth-Fixed (ECEF) coordinates~\cite{ecef}. 
Next, we perform a transformation from ECEF to East-North-Up (ENU) coordinates~\cite{enu}, ultimately establishing an OpenGL coordinate system with the scene center as the origin. 
In terms of camera trajectory selection, we prioritize retaining views from the Top-Pose and AdaLevel-Pose while selectively excluding trajectories from the Building-Pose to ensure the quality and geometric consistency for reconstruction.
For the InstantNGP training of each scene, we set training iterations to 18,000, which takes approximately 5 minutes on a single H20 GPU, achieving a good balance between computational efficiency and reconstruction quality.

\paragraph{Mesh Generation and Post-Processing.}
To derive 3D meshes from the outputs of InstantNGP, we employ the Marching Cubes algorithm within a 600-meter range at a voxel resolution of $512^3$.
Next, a series of refinements is incorporated to refine the mesh quality. 
We began by performing the connectivity analysis with Open3D~\cite{Zhou2018} to eliminate floating outliers.
Subsequently, we re-render the coarse mesh under each camera pose at a resolution of 360$\times$640 using NvDiff~\cite{Laine2020diffrast}, retaining only the vertices of the coarse mesh that were sampled during rendering. 
This step reduces the vertex count significantly, from approximately 600M to around 100M, thereby improving the processing efficiency for the subsequent stages.
To further enhance mesh quality, we apply PyMeshFix~\cite{pymeshfix} for topological repairs, filling in small holes. For semantic attribute integration, we select all 16 Top-Pose views and 54 highest AdaLevel-Pose views, resulting in a total of 72 observation viewpoints that are input into the AIE-SEG~\cite{xu2023analytical} model for semantic predictions. Additionally, we refine the mesh for water regions using the average height of water edge to land.
After reconstruction and refinement, we retain 88,256 scenes, ultimately filtering down to 50,028 scenes as the final dataset for Aerial-Earth3D, excluding those with excessive altitudes and overly smooth terrains. 




\subsection{Multi-View Score-Aggregation}\label{Score_Agg}
To fuse the \( N \) view 2D feature maps \( f_{n, i} \) (where \( i \in hw \) and \( n \in N \), representing either RGB images or flux features) into mesh vertex features \( \mathbf{F}_{j} \) (where \( j \in J \), and \( J \) is the number of mesh vertices), we follow a multi-step process:
We first compute the pixel view score \( S_{n, i} \), which is the cosine similarity between the mesh normal and the screen view direction.
The distance score \( D_{n, i} \) is calculated as:
\[D_{n, i} = 1 - \frac{d_{n, i}}{Z_{far}},\]
where \( d_{n, i} \) denotes the distance between pixel \( i \) in view \( n \) and the mesh surface, and \( Z_{far} \) represents the upper bound of the entire scene.
Subsequently,
we derive the rendering vertex ID map \( V_{n, i} \) (where \( V_{n, i} \in J \)) using the NvDiff rasterization operation.
Finally, the vertex features \( \mathbf{F}_{j} \) are formulated by combining the view score and distance score as follows:
\[
F_{j} = \frac{\sum^{C_j}_{c=1} f_{c} \cdot \psi(D_{c}, \tau_d) \cdot \psi(S_{c}, \tau_s)}{\eta},
\]
where \( C_j \) indicates the number of features with the same vertex index \( j \) across all view features \( f_{n, i} \), and
\[
\eta = \sum^{C_j}_{c=1} \psi(D_{c}, \tau_d) \cdot \psi(S_{c}, \tau_s).
\]
Here, \( \psi \) represents a power function, and both \( \tau_s \) and \( \tau_d \) are exponents that control the influence of the direction and distance of view, respectively.
Specifically, \( Z_{far} \) is set to 2.0, corresponding to 600 meters in the real scene, and both \( \tau_s \) and \( \tau_d \) are set to 3.0, indicating that the view distance and the view direction contribute equally to the aggregation of features.
Leveraging the efficient rasterization capabilities of NvDiff and PyTorch's index\_add($\cdot$) operation, we achieve efficient aggregation of multi-view features. The implementation is detailed in the following code snippet.

\subsection{Data Filter}
The utilization of diverse aerial data sources, such as official Google imagery and Airbus data within Google Earth Studio, combined with the presence of extreme building heights, has led to the reconstruction of scene meshes via Instant-NGP that exhibit irregular geometries. These irregularities primarily manifest in two issues: firstly, scene meshes reconstructed from high-altitude source aerial images, especially from Airbus, tend to overly smooth surfaces, and secondly, there are irregular geometrical distortions observed at the apexes of buildings.
To mitigate these issues, a scene height map-based filtering approach was implemented. Specifically, we employed Open3D to render 2D downward-looking height maps utilizing parallel downward-facing camera rays. A maximum height threshold \( t_h \) was applied to exclude scene meshes exceeding the specified height criterion. Following this, we employ height gradient analysis to filter out smooth meshes. This involves computing the height gradient map and excluding meshes that have an average gradient below a predefined minimum threshold \( t_g \).
Through this comprehensive filtering methodology, we successfully refined the dataset, resulting in a final collection of 50,028 scene meshes for Aerial-Earth3D.

\begin{figure*}
\centering
\includegraphics[width=0.85\linewidth]
{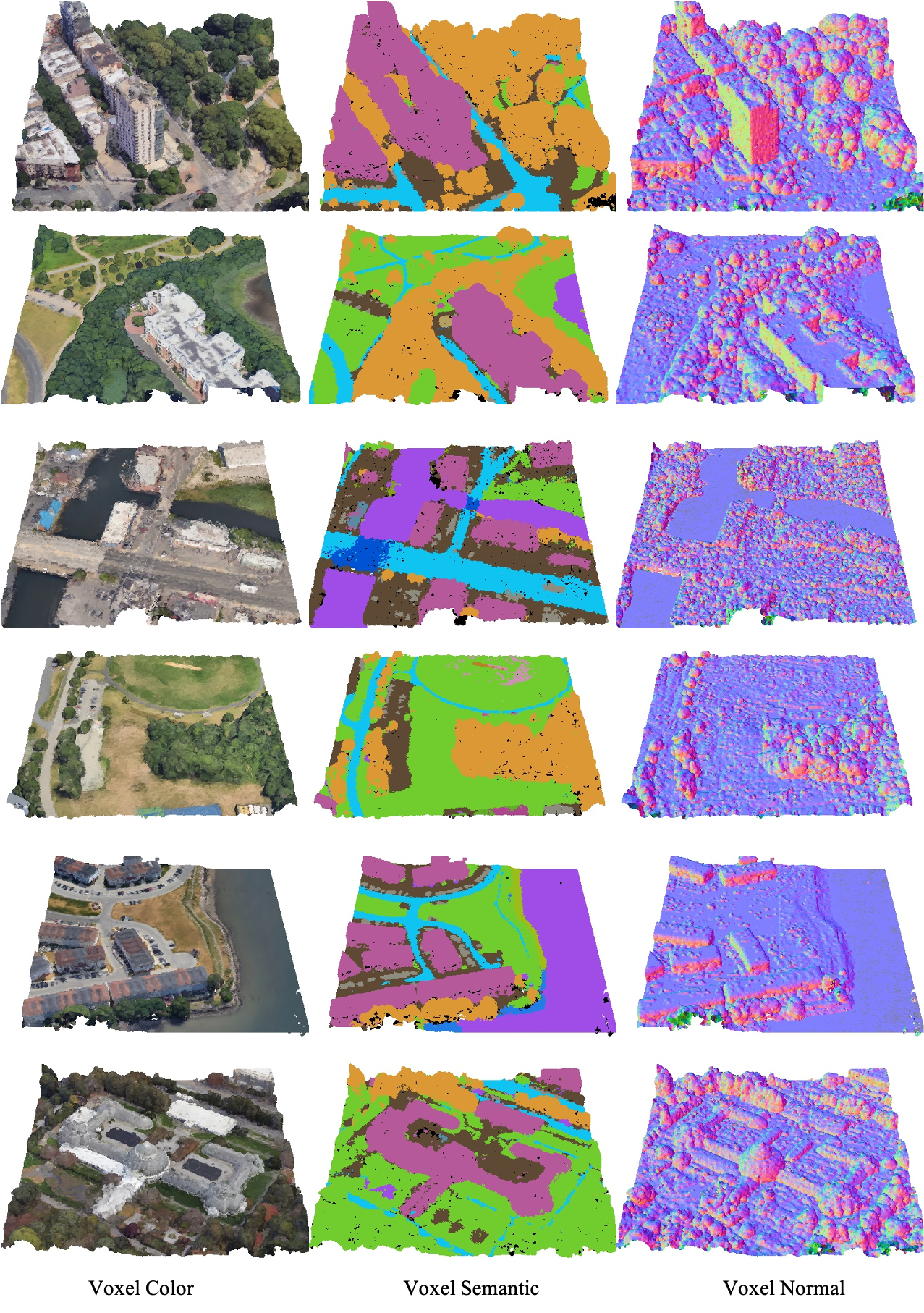}
\vspace{-0.2in}
   \caption{Voxel data visualizations of Aerial-Earth3D.}
   \label{fig:voxel_visual}
\vspace{-0.1in}
\end{figure*}

Figure~\ref{fig:voxel_visual} illustrates the voxel feature attributes, including color, semantics, and normals. The data preparation pipeline effectively reconstructs aerial scenes across diverse environments, capturing precise and varied 3D semantics as well as consistent normal directions. Furthermore, we believe that this dataset can significantly contribute to various tasks, such as 3D and 2D semantic segmentation, building height estimation, outline polygon prediction, and outdoor scene reconstruction for embedded AI and autonomous driving applications.

\subsection{Semantic Distribution}
\begin{figure}[t]
\centering
\includegraphics[width=0.46\textwidth]{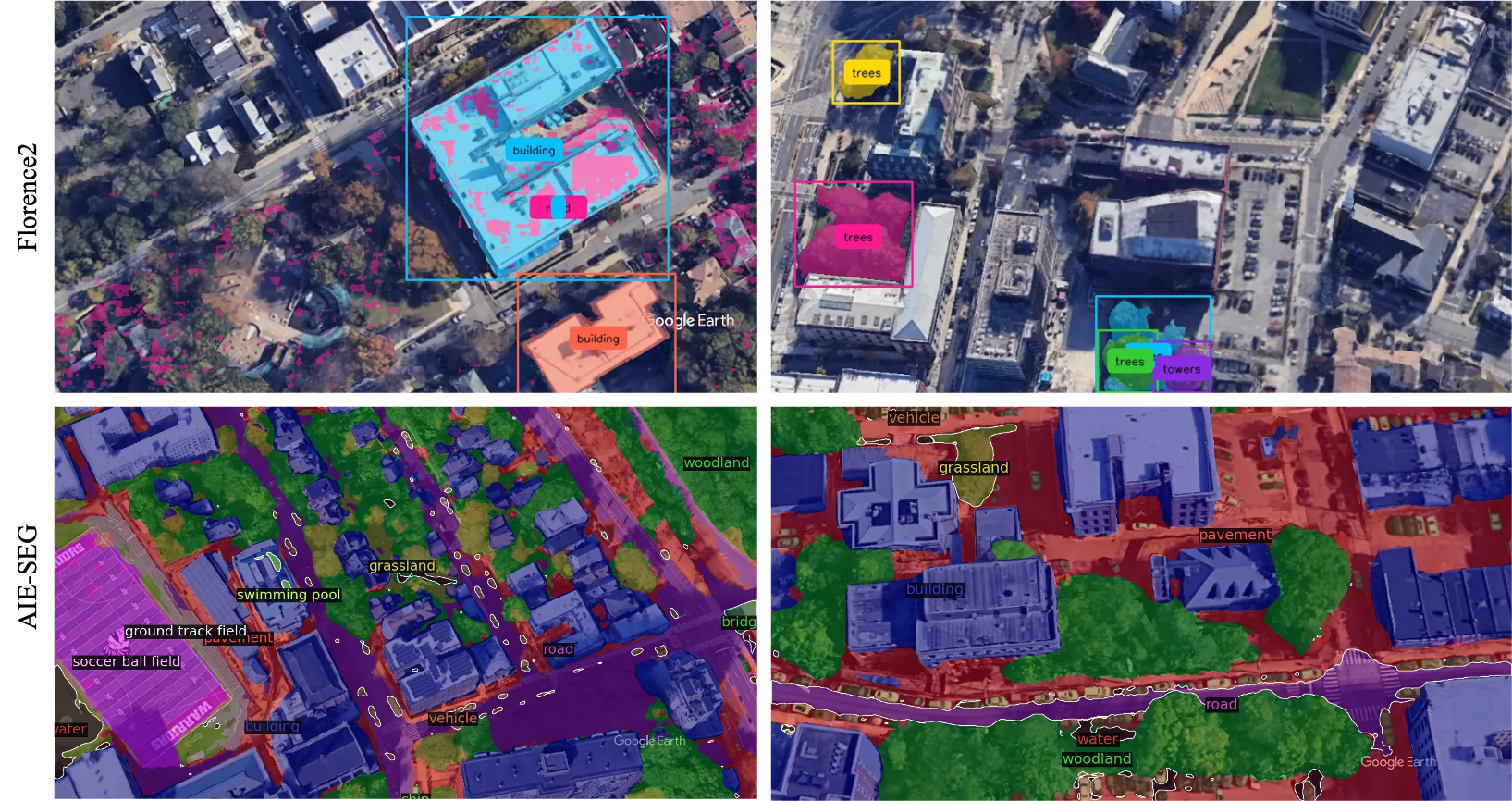}
\vspace{-0.1in}
\caption{Segmentation comparison between Florance2 and AIE-SEG.}
\vspace{-0.1in}
\label{fig:semantic_compare}
\end{figure}

To obtain the semantic attributes of scene meshes, we initially attempted to utilize semantic labels from the OpenStreetMap (OSM) dataset. However, the label tags in OSM lack standardization and contain numerous unlabeled regions, which impedes effective semantic annotation. Consequently, we reviewed publicly available segmentation models and selected Florence2~\cite{xiao2023florence2advancingunifiedrepresentation}, a state-of-the-art prompt-based vision foundation model primarily trained on natural images. 
As illustrated in the top row of Figure~\ref{fig:semantic_compare}, the Florence2 model only segments salient objects and fails to detect many targets in crowded aerial view images, such as buildings and trees. Additionally, Florence2 exhibits inefficiency, as it can predict only one prompt class per forward pass.
To address these limitations, we explored specialized models within the aerial domain and ultimately adopted the AIE-SEG aerial segmentation model~\cite{xu2023analytical}. AIE-SEG is trained on proprietary aerial data and supports 25 land cover classification types. The segmentation results are presented in the second row of Figure~\ref{fig:semantic_compare}. Compared to Florence2, AIE-SEG effectively segments both object classes (e.g., buildings, cars) and non-object classes (e.g., roads, trees, ground). Moreover, AIE-SEG is more efficient, capable of segmenting all 25 classes in a single forward pass.

For semantic integration, we selected a total of 72 observation viewpoints, comprising 16 Top-Pose views and 54 highest AdaLevel-Pose views. These viewpoints were input into the AIE-SEG model to generate semantic predictions. Subsequently, we integrated the semantic predictions into the scene meshes using the aggregation method.
The entire process takes approximately 2 minutes per scene.

\begin{figure}[ht]
\centering
\includegraphics[width=0.4\textwidth]{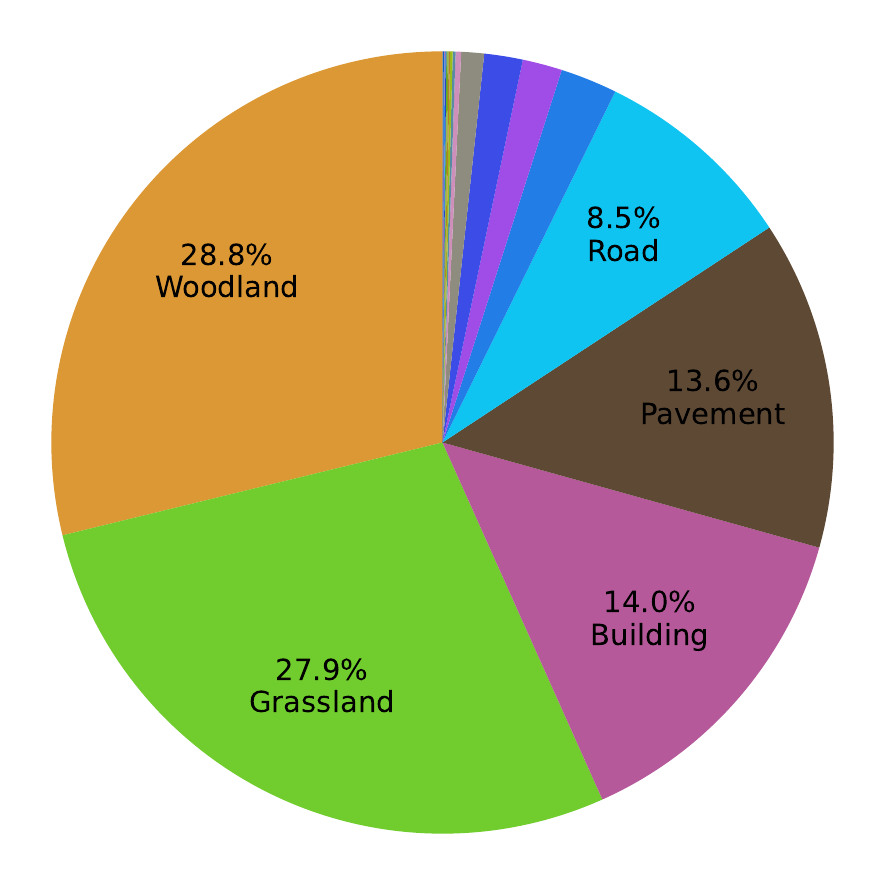}
\vspace{-0.1in}
\caption{Pie chart of the class distribution of Aerial-Earth3D.}
\vspace{-0.1in}
\label{fig:pie_semantic}
\end{figure}

Furthermore, we analyzed the semantic distribution of all scene meshes, as shown in Figure~\ref{fig:pie_semantic}. The analysis reveals that Woodland, Grassland, Building, Pavement, and Road are the top five classes. Detailed semantic distribution information is provided in Table~\ref{tab:semantic_classes_final}.

\begin{table}[h!]
\centering
\footnotesize
\renewcommand{\arraystretch}{1.2} 
\begin{tabular}{c|c|c|c}
\hline
\textbf{ID} & \textbf{Name}       & \textbf{Percentage (\%)} & \textbf{Color} \\ \hline
2   & Woodland             & 28.8137 & \cellcolor[RGB]{219,152,52}\textcolor{black}{[219, 152, 52]}  \\ \hline
3   & Grassland            & 27.8886 & \cellcolor[RGB]{113,204,46}\textcolor{black}{[113, 204, 46]}  \\ \hline
4   & Building             & 13.9675 & \cellcolor[RGB]{182,89,155}\textcolor{black}{[182, 89, 155]}  \\ \hline
9   & Pavement             & 13.5912 & \cellcolor[RGB]{94,73,52}\textcolor{black}{[94, 73, 52]}      \\ \hline
5   & Road                 & 8.4591  & \cellcolor[RGB]{15,196,241}\textcolor{black}{[15, 196, 241]}   \\ \hline
6   & Excavated Land       & 2.3485  & \cellcolor[RGB]{34,126,230}\textcolor{black}{[34, 126, 230]}   \\ \hline
8   & Water                & 1.6379  & \cellcolor[RGB]{160,76,231}\textcolor{black}{[160, 76, 231]}   \\ \hline
1   & Agriculture Field    & 1.5903  & \cellcolor[RGB]{60,76,231}\textcolor{black}{[60, 76, 231]}    \\ \hline
17  & Vehicle              & 0.9404  & \cellcolor[RGB]{141,140,127}\textcolor{black}{[141, 140, 127]} \\ \hline
21  & Soccer Ball Field    & 0.2278  & \cellcolor[RGB]{206,143,187}\textcolor{black}{[206, 143, 187]} \\ \hline
19  & Swimming Pool        & 0.1175  & \cellcolor[RGB]{89,140,163}\textcolor{black}{[89, 140, 163]}   \\ \hline
7   & Bare Land            & 0.1045  & \cellcolor[RGB]{156,188,26}\textcolor{black}{[156, 188, 26]}   \\ \hline
12  & Baseball Diamond     & 0.0610  & \cellcolor[RGB]{185,128,41}\textcolor{black}{[185, 128, 41]}   \\ \hline
10  & Ship                 & 0.0470  & \cellcolor[RGB]{133,160,22}\textcolor{black}{[133, 160, 22]}   \\ \hline
13  & Tennis Court         & 0.0463  & \cellcolor[RGB]{96,174,39}\textcolor{black}{[96, 174, 39]}    \\ \hline
11  & Storage Tank         & 0.0385  & \cellcolor[RGB]{43,57,192}\textcolor{black}{[43, 57, 192]}    \\ \hline
15  & Ground Track Field   & 0.0375  & \cellcolor[RGB]{18,156,243}\textcolor{black}{[18, 156, 243]}   \\ \hline
14  & Basketball Court     & 0.0290  & \cellcolor[RGB]{173,68,142}\textcolor{black}{[173, 68, 142]}   \\ \hline
16  & Bridge               & 0.0243  & \cellcolor[RGB]{0,84,211}\textcolor{black}{[0, 84, 211]}      \\ \hline
23  & Harbor               & 0.0141  & \cellcolor[RGB]{124,175,77}\textcolor{black}{[124, 175, 77]}   \\ \hline
20  & Roundabout           & 0.0080  & \cellcolor[RGB]{182,159,97}\textcolor{black}{[182, 159, 97]}   \\ \hline
24  & Greenhouse           & 0.0047  & \cellcolor[RGB]{46,58,176}\textcolor{black}{[46, 58, 176]}    \\ \hline
22  & Plane                & 0.0013  & \cellcolor[RGB]{43,147,240}\textcolor{black}{[43, 147, 240]}   \\ \hline
25  & Solar Panel          & 0.0012  & \cellcolor[RGB]{226,173,93}\textcolor{black}{[226, 173, 93]}   \\ \hline
18  & Helicopter           & 0.0001  & \cellcolor[RGB]{137,122,108}\textcolor{black}{[137, 122, 108]} \\ \hline
\end{tabular}
\caption{Semantic classes with ID, name, percentage, and color.}
\label{tab:semantic_classes_final}
\vspace{-0.2in}
\end{table}

\subsection{Train and Validation Dataset Split}
Given a filtered set of $50$k scene meshes, we generate $450$k voxel features, denoted as \( V_{\text{feat}} \). To construct a nearly uniformly distributed validation dataset, we first analyze the maximum voxel heights of each voxel feature. Subsequently, we uniformly partition the voxel features into $20$ subgroups based on their maximum voxel heights. From each subgroup, we randomly sample voxel features at a ratio of $1/120$ to form the validation set, ensuring that each subgroup contributes at least $8$ validation samples. This approach yields 447,000 training samples and 3,068 validation samples,
which is utilized to train and evaluate both StructVAE and TexVAE models.

For the flow matching training, to further mitigate the influence of smooth meshes, we exclude voxel features derived from Airbus data sources by employing an additional template-matching method. This method utilizes a mask template based on the Airbus watermark to exclude meshes reconstructed from Airbus aerial images, which produce 407,545 training samples and 2,801 validation samples. We use these data to train and evaluate ClassSFM, LatentSFM, and TexFM.

\section{Robust Generation}
The overall architecture of EarthCrafter consists of distinct structure and texture generation components that utilize three flow models and two variational autoencoder (VAE) models. During the inference process, we arranged these five models sequentially, such that the predictive output of each model serves as the input for the subsequent model. This sequential configuration, however, faces domain gap challenges, primarily stemming from inconsistencies between the outputs generated by earlier models and the inputs required for later models. For instance, we utilize the ground truth high-resolution voxel coordinates \( V_c \in \mathbb{R}^{L^3 \times 3} \) to train TexFM. However, the predicted voxel coordinates \( V'_c \in \mathbb{R}^{L^3 \times 3} \) may not align precisely with \( V_c \), leading to a degradation in performance.

To mitigate this discrepancy, we incorporate a range of voxel augmentations during model training, enhancing the model's robustness to variations in input data.

\subsection{Voxel Jagged Perturbation}
To address the discrepancies between StructFM and the texture models, we propose a voxel coordinate jagged perturbation augmentation policy specifically for training TexFM and TexVAE. Each voxel coordinate in \( V_c \in \mathbb{R}^{L^3 \times 3} \) is represented as \( p_c = (i, j, k) \) where \( i,j,k \in [0, L) \). 
The generation of jagged perturbation voxel coordinates \( V^J_c \in \mathbb{R}^{L^3 \times 3} \) involves the addition of random coordinate offsets to the source voxel coordinates \( p_c \). This is computed using the following formula:
\(
p'_c = (i', j', k'); \quad i' = i + \text{randint}(-1, 1); \quad j' = j + \text{randint}(-1, 1); \quad k' = k + \text{randint}(-1, 1).
\)
Subsequently, we apply a unique operation to remove any duplicated jagged coordinates, and the resulting clean jagged voxel coordinates are utilized as training inputs. For the jagged perturbation of the textural latent variable \( T_{La} \in \mathbb{R}^{L^3 \times c_t} \), we modify only the voxel coordinates, preserving the voxel feature values.

\begin{figure*}
\centering
\includegraphics[width=0.95\linewidth]
{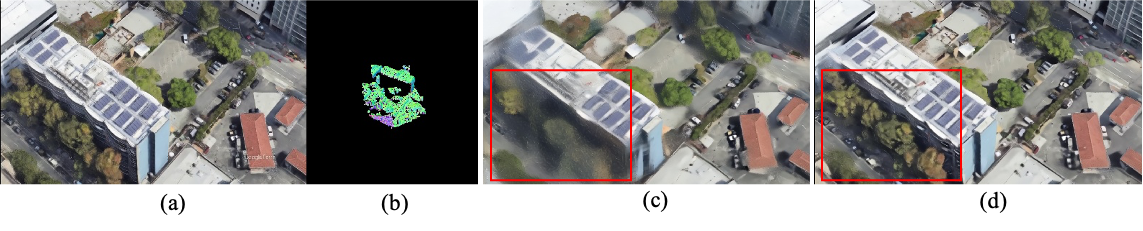}
\vspace{-0.1in}
\caption{Results of appearance changing caused by jagged perturbation.
 (b) means conditional voxels from the conditional image's depth. ; (d) denotes generated appearance by using jagged perturbation.}
 \label{fig:jagged_per}
\vspace{-0.1in}
\end{figure*}

The results presented in Table~\ref{tab:global_texvae_jagged} and Table~\ref{tab:global_texfm_jagged} indicate that the Voxel Jagged Perturbation may initially degrade performance during single model evaluations. However, in the context of sequential inference, we observe that models trained with jagged perturbation exhibit a more photorealistic output, as depicted in Figure~\ref{fig:jagged_per}.

\begin{table}[t!]
  \scriptsize
  \centering
  \vspace{-0.1in}
  \setlength{\tabcolsep}{3.5pt} 
    \begin{tabular}{c|c|c|c|c|c}
    \hline 
    Method & jagged  perturbation & PSNR{\small{}$\uparrow$} & L1{\small{}$\downarrow$} & LPIPS{\small{}$\downarrow$} & SSIM{\small{}$\downarrow$}\tabularnewline
    \hline
     TexVAE & & \textbf{22.9} & \textbf{0.040} & \textbf{0.280}  & \textbf{0.301} \tabularnewline
     TexVAE$^{+}$ & $\checkmark$ & 22.5 & 0.041 & 0.292  & 0.314 \tabularnewline
    \hline 
    \end{tabular}
  \caption{TexVAE results on global data with/w.o. jagged perturbation.}
  \label{tab:global_texvae_jagged}
\end{table}

\begin{table}[t!]
  \scriptsize
  \centering
  \vspace{-0.1in}
  \setlength{\tabcolsep}{3.0pt} 
    \begin{tabular}{c|c|c|c|c}
    \hline 
    Method & jagged  perturbation & $PSNR_{img}${\small{}$\uparrow$} & $FID_{uncond}${\small{}$\downarrow$} & $FID_{sem}${\small{}$\downarrow$} \tabularnewline
    \hline
     TexFM & & \textbf{20.6} & \textbf{21.2} & \textbf{36.7} \tabularnewline
     TexFM$^{+}$ & $\checkmark$ & 20.3 & 25.3  & 40.1 \tabularnewline
    \hline 
    \end{tabular}
  \caption{Results of TexFM on global dataset. $PSNR_{img}$ denotes PSNR metric under image condition, $FID_{uncond}$ and $FID_{sem}$ represent FID metric under empty condition and semantic map separately.}
  \label{tab:global_texfm_jagged}
    \vspace{-0.1in}
\end{table}

\subsection{Voxel Roughening}

To address the discrepancy between the ClassSFM (the first stage in StructFM) and the LatentSFM (the second stage in StructFM), we implement a voxel roughening augmentation during the training of LatentSFM. Voxel roughening comprises two primary operations: voxel dilation and voxel simplification.

\begin{figure}
\centering
\includegraphics[width=0.46\textwidth]{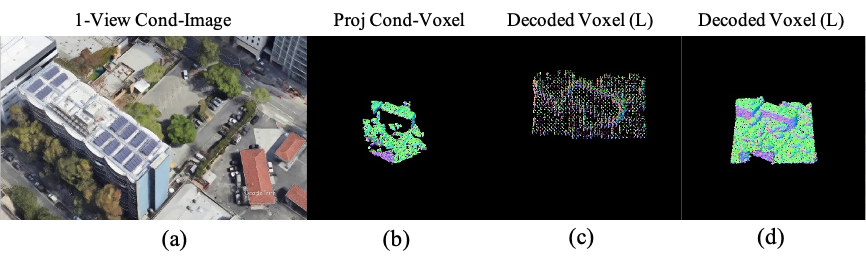}
\vspace{-0.1in}
\caption{Visualization comparison with and without roughening. (c) means without applying roughening, (d) means using voxel roughening.
}
\label{fig:compare_roughening}
\vspace{-0.1in}
\end{figure}

We begin by feeding the voxel coordinates \( V_c \in \mathbb{R}^{L^3 \times 3} \) into the StructVAE Encoder to generate the ground truth structural latents \( S_{La} \in \mathbb{R}^{\left(\frac{L}{8}\right)^{3} \times c_s} \). The latent voxel coordinates in \( S_{La} \) are represented as \( V_{La} \in \mathbb{R}^{\left(\frac{L}{8}\right)^{3} \times 3} \).
Then, we apply a dilation operation with a kernel size of \( d_l \) to expand the sparse latent voxels \( V_{La} \). Following the dilation, we utilize the voxel simplification operation, which involves downsampling the dilated voxels and then upsampling them by a factor of \( s_l \). This process ultimately yields the roughened latent voxel coordinates \( V^R_{La} \in \mathbb{R}^{\left(\frac{L}{8}\right)^{3} \times 3} \).
Subsequently, we derive the roughened structural latents \( S^R_{La} \in \mathbb{R}^{\left(\frac{L}{8}\right)^{3} \times c_s} \) by setting the latent features of voxels that do not exist in \( S_{La} \) to zero, while preserving the original feature values for voxels present in \( S_{La} \). The roughened voxel coordinates \( V^R_{La} \) are then utilized as the training inputs for LatentSFM.
During the inference phase, we employ the voxel roughening operation on the sparse voxel outputs generated by ClassSFM. This process is designed to align these outputs more closely with the training inputs utilized by LatentSFM. As illustrated in Figure~\ref{fig:compare_roughening}, voxel roughening induces significant changes. A comparative analysis of Figure~\ref{fig:compare_roughening}(c) and Figure~\ref{fig:compare_roughening}(d) reveals that, in the absence of roughening alignment, the discrepancies between the generated coarse voxels and the target voxels can lead to failure in the generation process.

\subsection{Voxel Normal Drop}

\begin{figure*}
\centering
\includegraphics[width=0.95\linewidth]
{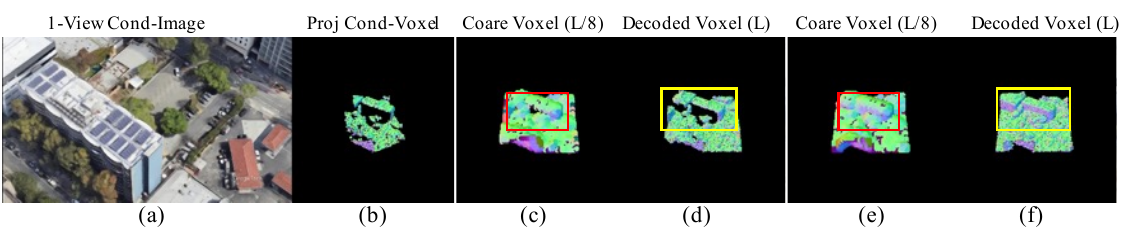}
\vspace{-0.1in}
   \caption{Results of geometry changing caused by normal drop augmentation.
     (b) means conditional voxels from the conditional image's depth; (e) and (f) represent generated voxels after applying normal drop.}
     \label{fig:normal_aug_change}
\vspace{-0.1in}
\end{figure*}

The limitations of aerial views result in unseen regions within the scene mesh, particularly in crowded building environments. Additionally, the Instant-NGP algorithm may encounter difficulties with semi-transparent surfaces, such as water and glass curtain walls of buildings. These two issues contribute to a low ratio of scene meshes that contain holes. While small holes in the scene mesh can be filled using PyMeshFix, larger holes are not effectively addressed by third-party tools. Consequently, these holes lead to training voxels that incorporate voids, resulting in both hole-condition voxels and hole-target voxels.
As we support one-view image condition generation, and we produce condition voxels by projecting image depth into 3D space and voxelizing it, as shown in Figure~\ref{fig:normal_aug_change}(b), which provides only partial information. This limitation increases the likelihood of the generated scene voxels exhibiting holes.
To resolve this issue, we propose a normal direction-based conditional voxel dropping augmentation. Empirically, we have observed that holes in the scene mesh often appear on the side surfaces of objects, such as building facades and the sides of trees. Hence, we drop conditional voxels based on their normal directions.

\begin{figure}[t!]
\centering
\includegraphics[width=0.46\textwidth]{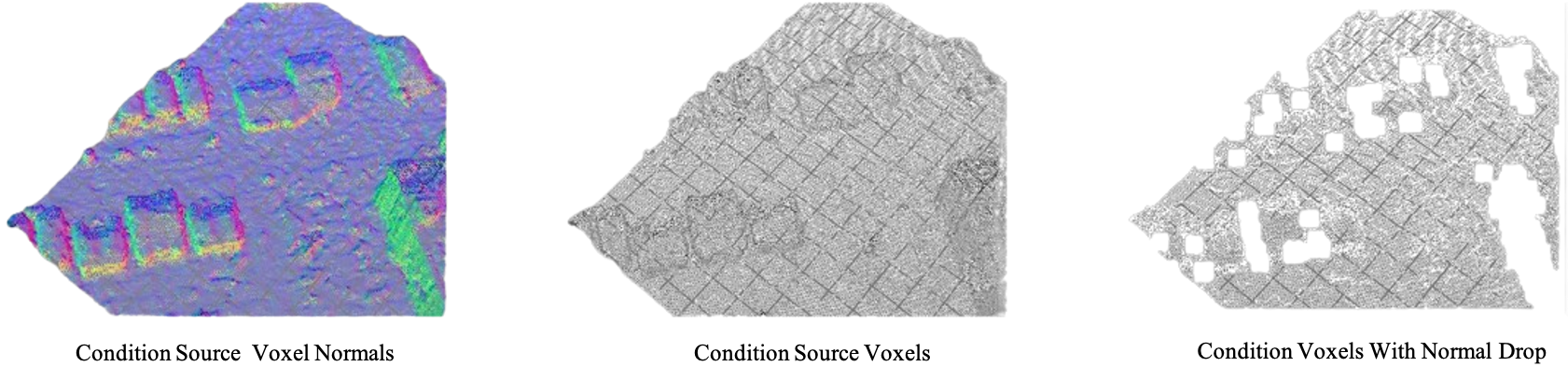}
\vspace{-0.1in}
\caption{Visualization of condition voxels with voxel normal drop.}
\label{fig:normal_drop_visual}
\vspace{-0.1in}
\end{figure}

In detail, we first extract two primary normal directions, \( N_x \) and \( N_y \), from the conditional voxel normals \( N_C \) in the XY-plane. Next, we randomly select one normal direction \( N_s \) from \( N_x \) and \( N_y \) and introduce a small amount of random noise to \( N_s \). We then compute the cosine similarity \( S_n \) between \( N_s \) and \( N_C \). Subsequently, we drop voxels where \( S_n > \sigma_s \), with \( \sigma_s \) being the dropping score threshold. Finally, we apply dilation and erosion operations to ensure that the resulting drop regions are smoother and more connected. Figure~\ref{fig:normal_drop_visual} illustrates the outcomes of the voxel normal drop process, showcasing the removal of voxels directed towards the left.

Similar to the effectiveness of Voxel Jagged Perturbation, the results presented in Table~\ref{tab:ablation_structfm2} indicate that applying normal drop augmentation degrades performance during single-task evaluations. This occurs because the predicted voxel scenes are complete, while the ground truth voxels may contain holes. However, the results illustrated in Figure~\ref{fig:normal_aug_change} display the geometric changes after applying the voxel normal drop augmentation. When comparing Figure~\ref{fig:normal_aug_change}(c) with Figure~\ref{fig:normal_aug_change}(e) and Figure~\ref{fig:normal_aug_change}(d) with Figure~\ref{fig:normal_aug_change}(f), it is evident that even with incomplete and irregular conditional voxels, as seen in Figure~\ref{fig:normal_aug_change}(b), both structure flow models (ClassSFM and LatentSFM) can successfully generate complete scene voxels.

\begin{table}
  \scriptsize
  \centering
  \vspace{-0.1in}
    \begin{tabular}{c|c|c|c|c|c}
    \hline 
    Method & S-Num & S-Index & NormalCrop & $mIoU^3${\small{}$\uparrow$}  & $mIoU^0${\small{}$\uparrow$} \tabularnewline
    \hline
     ClassSFM & 2 & 1 &  & 84.7 & - \tabularnewline
     LatentSFM & 2 & 2 &  & \textbf{86.1} & \textbf{25.4} \tabularnewline
    \hline 
     ClassSFM$^{\dagger}$ & 2 & 1 & $\checkmark$ & 83.9 & -\tabularnewline
     LatentSFM$^{\dagger}$ & 2 & 2 & $\checkmark$ & \textbf{84.3} & 23.3 \tabularnewline
    \hline
    \end{tabular}
  \caption{Results of StructFlows on train data under image condition. S-Num: total stage number; S-Index: stage index; NormalDrop: condition drop augmentation based on condition normal direction; $mIoU^3$ and $mIoU^0$ denote voxel structure metric at $L/8$ level and $L$ level.}
  \label{tab:ablation_structfm2}
\end{table}

\subsection{Unbalance Sampling}
As we sample interesting scene sites from the mainland of the United States, we encounter a significant class imbalance problem. Table~\ref{tab:semantic_classes_final} illustrates the percentage of each class, 
revealing that the classes are dominated by ground classes, such as Woodland, Grassland, pavement, and road, the building class constitutes only $14\%$ of the total dataset.
To address this imbalance issue, we implement a weighted sampling policy that emphasizes the maximum scene voxel height, denoted as $Z_i$. The rationale behind this approach is that a larger $Z_i$ indicates a scene with a higher density of buildings. The sampling weight $W_i$ is calculated using the following equations:
\begin{align}
W_i & = \frac{z_i}{\sum_{i} z_i}, z_i = (\frac{\text{min}(Z_i, 200)}{10})^\alpha 
\end{align}

\section{Limitations}
While EarthCrafter exhibits enhanced and diverse capabilities for generating 3D scenes, However it does have certain limitations.
First, the limited number of aerial images available for each site results in suboptimal scene mesh geometry generated by Instant-NGP. This limitation manifests in several ways: 1) noticeable holes and distortions can be observed at the corners of trees and the upper portions of tall buildings, and 2) the geometry of roofs and ground surfaces tends to possess higher fidelity compared to side surfaces, such as building façades. These deficiencies negatively influence the overall generation quality, particularly affecting the representation of high-rise structures. Therefore, an enhancement process for the scene mesh geometry may be warranted.
Second, the EarthCrafter pipeline is relatively lengthy, comprising five models: three flow models and two variational autoencoders (VAEs). This complexity may hinder efficiency. Additionally, the domain gap between the outputs of the previous stages and the training inputs further degrades overall performance, suggesting that simpler approaches warrant exploration to streamline the pipeline.
Thirdly, to align with input conditions, our method requires intricate 3D and 2D mapping operations, which constrain flexibility and applicability in out-of-domain scenarios. There is a pressing need to investigate more adaptable methods for condition injection, such as leveraging text and image embeddings that do not necessitate strict 3D alignment. This would enhance the system's versatility for use in a wider array of applications.
Finally, since we exclusively utilize bird’s eye view (BEV) aerial images from Google Earth Studio, the quality of scene generation in first-person view (FPV) mode is suboptimal. Integrating FPV data in the same location, such as street view images, could improve rendering quality in FPV mode.

\clearpage
\onecolumn
\begin{lstlisting}[language=Python, caption={\textbf{3D Texture Fusion Process Implementation}}.]
# planar_vertices represents the mesh vertices, planar_faces represents the mesh faces
# N is the number of views

feature = torch.zeros((planar_vertices.shape[0], 3), dtype=torch.float32).to(device)
weights = torch.zeros((planar_vertices.shape[0], 1), dtype=torch.float32).to(device)

for i in range(N):
    
    cur_image = load_from_path_i
    cur_c2w = load_from_pose_i

    cur_render_rast, cur_depth = NvDiff.render_scene_mesh(cur_c2w, planar_vertices, planar_faces)

    cur_gb_pos, _ = NvDiff.interpolate_one(planar_vertices, cur_render_rast, planar_faces)

    cur_pix_to_face = cur_render_rast[..., 3:4].long().squeeze() - 1
    cur_valid_render_mask = cur_pix_to_face > -1
    cur_valid_mask = torch.logical_and(cur_valid_render_mask, water_mask)

    cur_valid_latents = cur_image[cur_valid_mask]
    cur_valid_face_idxs = cur_pix_to_face[cur_valid_mask]
    cur_valid_vertex_idxs = planar_faces[cur_valid_face_idxs]

    cur_gb_pos = cur_gb_pos[0, cur_valid_mask]
    cur_t = torch.from_numpy(cur_c2w[:3, 3]).to(cur_gb_pos.device).reshape((1, 3)).float()
    cur_gb_viewdirs = F.normalize(cur_t - cur_gb_pos, dim=-1)
    cur_valid_normals = face_normals[cur_valid_face_idxs]
    cur_valid_normals = F.normalize(cur_valid_normals, p=2, dim=-1, eps=1e-6)
    cur_view_score = (cur_gb_viewdirs * cur_valid_normals).sum(dim=1).abs()
    cur_view_score = torch.pow(cur_view_score, exponent_view)

    cur_valid_depth = torch.clamp(cur_depth[0, cur_valid_mask], 0, 2.0)
    cur_depth_score = 1.0 - cur_valid_depth / 2.0
    cur_depth_score = torch.clamp(cur_depth_score, 0, 1.0)
    cur_depth_score = torch.pow(cur_depth_score, exponent_depth)

    cur_score = (cur_view_score * cur_depth_score).view(-1, 1, 1).repeat(1, 3, 1).view(-1, 1)

    cur_valid_vertex_idxs = cur_valid_vertex_idxs.flatten()
    cur_valid_latents = cur_latents.view(-1, latent_channel)

    cur_valid_latents = cur_valid_latents.unsqueeze(1).repeat(1, 3, 1).view(-1, latent_channel)
    cur_valid_latents = cur_valid_latents * cur_score

    feature.index_add_(0, cur_valid_vertex_idxs, cur_valid_latents)
    weights.index_add_(0, cur_valid_vertex_idxs, cur_score)

feature = feature / (weights + 1e-6)
\end{lstlisting}
\twocolumn

\bibliography{aaai2026}

\end{document}